\documentclass[twocolumn, 10pt, compsoc]{IEEEtran}

\pdfoutput=1
\usepackage{preamble} 
\usepackage{flushend}
\usepackage{fancyhdr}
\usepackage{txfonts}

\edef\keptrmdefault{\rmdefault}
\edef\keptsfdefault{\sfdefault}
\edef\keptttdefault{\ttdefault}

\usepackage[OT1]{fontenc}
\usepackage{concmath}
\edef\rmdefault{\keptrmdefault}
\edef\sfdefault{\keptsfdefault}
\edef\ttdefault{\keptttdefault}
\title{Data Smashing 2.0: \\Sequence Likelihood (SL) Divergence For Fast Time Series Comparison}
\author{

    \IEEEauthorblockN{
        Yi Huang, 
        Ishanu Chattopadhyay
    }\\
    \IEEEauthorblockA{
        Institute of Genomics and Systems Biology and\\ 
        Department of Medicine, University of Chicago, Chicago, IL, 60637, USA\\
        \{yhuang10, ishanu\}@uchicago.edu
    }
}

\begin{document}  
\maketitle

\begin{abstract}
Recognizing subtle historical patterns is central to modeling and forecasting problems in time series analysis. Here we introduce and develop a new approach to quantify deviations in the underlying hidden generators of observed data streams, resulting in a new efficiently computable universal metric for time series. The proposed metric is \textit{universal} in the sense that we can compare and contrast data streams regardless of where and how they are generated, and without any feature engineering step. The approach proposed in this paper is conceptually distinct from our previous work on data smashing~\cite{chattopadhyay2014data}, and vastly improves discrimination performance and computing speed. The core idea here is the generalization of the notion of KL divergence often used to compare probability distributions to a notion of divergence in time series. We call this generalization the sequence likelihood (SL) divergence and show that it can be used to measure deviations within a well-defined class of discrete-valued stochastic processes. We devise efficient estimators of SL divergence from finite sample paths, and subsequently formulate a universal metric useful for computing distance between  time series produced by hidden stochastic generators. We illustrate the superior performance of the new \textsf{smash2.0} metric with synthetic data against the original data smashing algorithm and dynamic time warping (DTW)~\cite{petitjean2011global}. Pattern disambiguation in two distinct applications involving electroencephalogram data and gait recognition is also illustrated. We are hopeful that the \textsf{smash2.0} metric introduced here will become an important component of the standard toolbox used in classification, clustering and inference problems in time series analysis.
\end{abstract}


\allowdisplaybreaks
\section{Introduction}
Efficiently learning stochastic processes is a key challenge in analyzing  time-dependency in domains where randomness cannot be ignored. For such learning to occur, we need to define either a distance metric or, more generally, a measurement of similarity to compare and contrast time series. Examples of such similarity measurement from the literature include the classical $l_p$ distances and $l_p$ distances with dimensionality reduction \cite{lin2003symbolic}, the short time series distance (STS)\cite{moller2003fuzzy}, which takes into account of irregularity in sampling rates, the edit based distances\cite{navarro2001guided} with generalizations to continuous sequences\cite{chen2005robust}, and the dynamic time warping (DTW)\cite{petitjean2011global}, which is used extensively in the speech recognition community. However these measurement of similarity all have either one or both of the following limitations. First, dimensionality reduction and feature selection heavily relies on domain knowledge and inevitably incurs trade-off between precision and computability. Most importantly, it necessitates the attention of human experts and data scientists. Secondly, when dealing with data from non-trivial stochastic process dynamics, state of the art techniques might fail to correctly estimate the similarity or lack thereof between exemplars. For example, suppose two sequences recording $n$ tosses of a fair coins, use $1$ to represent a head and $0$, tail. The two sequences are extremely unlikely to share any similarity on the face value, $i.e.$ they have a large pointwise distance, but they are generated by the same process. A good measurement of similarity should strive to disambiguate the underlying processes. The \textsf{Smash2.0} metric introduced here addresses both these limitations.

When presented with finite sample paths, the \textsf{Smash2.0} algorithm is specifically designed to estimate a distance between the generating models of the time series samples.  The intuition for the \textsf{Smash2.0} metric  follows from a basic result in information theory: If we know the true distribution $p$ of the random variable, we could construct a code with average description length $H(p)$, where $H(\cdot)$ is the entropy of a distribution. If, instead, we used the code for a distribution $q$, we would need $H(p) + D(p||q)$ bits on the average to describe the random variable. Thus, deviation in the distributions show up as KL divergence. If we can generalize the notion of KL divergence to processes, then it might be possible to quantify deviations in process dynamics via an increase in the entropy rate by the corresponding divergence. 

Our ultimate goal is to design an algorithm that operates on a pair of data streams taking values in a finite alphabet. Nevertheless, to establish the correctness of our algorithm, we need to decide on a specific scheme for representing stochastic processes taking values in the alphabet. We further assume that our processes are ergodic and stationary. The specific modeling paradigm for modeling stochastic processes we use in this paper is called Probabilistic Finite-State Automaton, or PFSA for short, which has been studied in \cite{crutchfield1994calculi,dupont2005links,chattopadhyay2014data,chattopadhyay2014causality}. PFSA can model discrete-valued stochastic processes that are not Markov of any finite order\cite{ching2006markov}. It is also shown in \cite{dupont2005links} to be able to approximate any hidden Markov model (HMM) with arbitrary accuracy. Moreover, PFSA has the property that many key statistical quantities of the processes they generate, such as entropy rate\cite{cover2012elements} and KL-divergence\cite{matthews2016sparse}, have closed-form formulae. Here we want to point out the resemblance of the PFSA model to the variational autoencoder (VAE) \cite{rezende2014stochastic,kingma2013auto} framework. The inference of PFSA from the input can be thought as the training of the encoder in a VAE, and the performance of both the VAE and the PFSA model are evaluated by the log-likelihood of input as being generated by the inferred models.

The work that has inspired the development of \textsf{Smash2.0} is the data smashing algorithm (\textsf{Smash}) proposed in \cite{chattopadhyay2014data}. \textsf{Smash} is also based on PFSA modeling and designed directly to represent the similarity between the generating models rather than sample paths. However, as while as both \textsf{Smash} and \textsf{Smash2.0} have the advantage of not requiring dimensionality reduction or domain knowledge for feature extraction, \textsf{Smash2.0} is much more computationally efficient than \textsf{Smash}. 

The remaining of the paper is organized as follows. In Sec.~\ref{sec:Foundation}, we introduce basic concepts of stochastic processes and establish the correspondence between processes and labeled directed graphs via the core concept of causal state. The definition and basic properties of PFSA are introduced by the end of Sec.~\ref{subsec:FromQToPA}. In Sec.~\ref{sec:PFSAGenerator}, we answer the question of when a stochastic process has a PFSA generator. An inference algorithm, \algo, of PFSA is given in Sec.~\ref{sec:inference_PFSA}. In Sec.~\ref{subsec:Entropies}, we introduce the notion of irreducibility of PFSA and the closed-form formulae for entropy rate and KL divergence of the processes generated by irreducible PFSA. We conclude the section with log-likelihood convergence. In Sec.~\ref{sec:Smash2.0} we introduce the definition of \textsf{Smash2.0} together with quantization of continuous sequences. The comparison of \textsf{Smash2.0} to \textsf{Smash} and \textsf{fastDTW} is given in Sec.~\ref{subsec:Compare_SmashsAndFastDTW}. In Sec.~\ref{sec:Application}, we apply \textsf{Smash2.0} to two real world problems.

\section{Foundation}
\label{sec:Foundation}
\subsection{Stochastic Processes and Causal States}
\label{subsec:StochasticProcAndCausalStates}
In this paper we study the generative model for stationary ergodic stochastic processes~\cite{peebles2001probability,crutchfield1994calculi} over a finite alphabet. Specifically, we consider a set of $\Sigma$-valued random variables $\set{X_t}_{t\in\mathbb{N}^{+}}$ indexed by positive integers representing time steps. By stationary, we mean strictly stationary, \ie the finite-dimensional distributions~\cite{doob1990stochastic} are invariant of time. By ergodic, we mean that all finite-dimensional distributions can be approximated with arbitrary accuracy with long enough realization. We are especially interested in processes in which the $X_i$s are \emph{not} independent. 

We denote the alphabet by $\Sigma$ and use lower case Greeks (e.g.~$\sigma$ or $\tau$) for symbols in $\Sigma$. We use lower case Latins (e.g.~$x$ or $y$) to denote sequences of symbols, $x=\sigma_1\sigma_2\dots\sigma_n$ for example, with the empty sequence denoted by $\lambda$. The length of a sequence $x$ is denoted by $|x|$. The set of sequences of length $d$ is denoted by $\Sigma^{d}$, and the collection of sequences of finite length is denoted by $\Sigma^{\star}$, \ie $\Sigma^{\star}= \bigcup_{d=0}^{\infty}\Sigma^{d}$. We use $\Sigma^{\omega}$ to denote the set of infinitely long sequences, and $x\Sigma^{\omega}$ to denote the collection of infinite sequences with $x\in\Sigma^{\star}$ as prefix. We note that, since all sequences can be viewed as prefixed by $\lambda$, we have $\lambda\Sigma^{\omega}=\Sigma^{\omega}$.

We note that $\SR = \set{x\Sigma^{\omega}\left| x \in \Sigma^{\star}\right.}$ is a semiring over $\Sigma^{\omega}$. Let $Pr\paren{X_1\dots X_n=\sigma_1\dots\sigma_n}$ denote the probability of the process producing a realization with $X_i = \sigma_i$ for $i = 1, \dots, n$, it is straightforward to verify that $\mu:\SR\rightarrow [0, 1]$ defined by  
\begin{equation}
\label{eq:DefinitionOfPremeasure}
    \mu\paren{\sigma_1\dots\sigma_n\Sigma^{\omega}} = Pr\paren{X_1\dots X_n=\sigma_1\dots\sigma_n},
\end{equation}
is a premeasure on $\SR$. By Charath\'eodory extension theorem, the $\sigma$-finite premeasure $\mu$ can be extended uniquely to a measure over $\F=\sigma(\SR)$, where $\sigma(\SR)$ is the $\sigma$-field generated by $\SR$. Denoting the measure also by $\mu$, we see that every stochastic process induces a probability space $\paren{\Sigma^{\omega}, \F, \mu}$ over $\Sigma^{\omega}$. In light of Eq.~\eqref{eq:DefinitionOfPremeasure} and also for notational brevity, we denote $\mu\paren{x\Sigma^{\omega}}$ by $Pr(x)$ when no confusion arises. We note that $Pr(\lambda) = \mu\paren{\lambda\Sigma^{\omega}} = \mu\paren{\Sigma^{\omega}} = 1$. We refer to Chap.~1 of \cite{klenke2013probability} as a more formal introduction to the measure-theory knowledge used here.

Taking one step further, and denoting the collection of all measures over $\paren{\Sigma^{\omega}, \mathcal{F}}$ by $\mathcal{M}_{\Sigma}$, we see that we can get a family of measures in $\mathcal{M}_{\Sigma}$ from a process in addition to $\mu$.

\begin{defn}[Observation Induced Measures]
For an observed sequence $x\in\Sigma^{\star}$ with $Pr(x)> 0$, the measure $\mu_x$ is the extension to $\F$ of the premeasure defined on the semiring $\SR$ given by
\begin{equation*}
    \mu_x\paren{y\Sigma^{\omega}} = \frac{Pr\paren{xy}}{Pr(x)},
\end{equation*}
for all $y\in\Sigma^{\star}$.
\end{defn}

Now we introduce the concept of Probabilistic Nerode Equivalence, which was first introduced in \cite{chattopadhyay2008structural}. 
\begin{defn}[Probabilistic Nerode Equivalence]
For any pair of sequences $x, y\in\Sigma^{\star}$, $x$ is equivalent to $y$, written as $x\sim y$, if and only if either $Pr(x) = Pr(y) = 0$, or $\mu_x = \mu_y$. 
\end{defn}
One can verify that the relation defined above is indeed an equivalence relation which is also \emph{right-invariant} in the sense that $x\sim y \Rightarrow xz\sim yz$, for all $z\in\Sigma^{\star}$. We denote the equivalence class of sequence $x$ by $[x]$. We note that $\mu_{[x]}$ is well-defined because $\mu_x=\mu_y$ for $x, y\in[x]$. An equivalence class is also called a \textbf{causal state}~\cite{chattopadhyay2014data} since the distribution of future events preceded by possibly distinct $x,y\in[x]$ are both determined by $\mu_{[x]}$. We denote $\mu_{[x]}\paren{y\Sigma^{\omega}}$ by $Pr_{[x]}(y)$ when no confusion arises. We note that $\mu_{[\lambda]} = \mu$.

\begin{rem}
Since the equivalence class $\set{x\in\Sigma^{\star}|Pr(x) = 0}$ plays no role in our future discussion, we ignore it as a causal state from this point on.
\end{rem}

\begin{defn}[Derivatives]
\label{def:Derivatives}
For any $d\in\mathbb{N}^{+}$, the \textbf{$d$-th order derivative} of an equivalence class $[x]$, written as $\phi^{d}_{[x]}$, is defined to be the marginal distribution of $\mu_{[x]}$ on $\Sigma^d$, with the entry indexed by $y$ denoted by $\phi^{d}_{[x]}(y)$. The first-order derivative is also called the \textbf{symbolic derivative} in \cite{chattopadhyay2014data} since $\Sigma^{1} = \Sigma$, and is denoted by $\phi_{[x]}$ for short. The derivative of a sequence is that of its equivalence class, \ie $\phi^{d}_{x} = \phi^{d}_{[x]}$. We note that $\phi^{d}_{\lambda}$ is the marginal distribution of $\mu$ on $\Sigma^{d}$, and is denoted by $\phi^{d}$ for short.
\end{defn}

\subsection{From Causal States to Probabilistic Automaton}
\label{subsec:FromQToPA}
From now on, we denote the set of causal states of a process by $Q$ when no confusion arises. We start this section by showing that there is a labeled directed graph~\cite{bondy2008graph} associated with any stochastic process. 

For any $q\in Q$ and $\sigma\in\Sigma$ such that $Pr_{q}(\sigma) > 0$, by right-invariance of probabilistic Nerode equivalence, there exists a $q'\in Q$, such that $x\sigma\in q'$ for all $x\in q$. Whenever the scenario described happens, we can put a directed edge from $q$ to $q'$ and label it by $\sigma$ and $Pr_{q}(\sigma)$, and by doing this for all $q\in Q$ and $\sigma\in\Sigma$, we get a (possibly infinite) labeled directed graph with vertex set $Q$. 

\begin{exmpl}[An Order-One Markov Process]
\label{exmpl:M2}
We now carry out the construction described above on an order-$1$ Markov process~\cite{gagniuc2017markov} over alphabet $\Sigma = \set{0, 1}$, in which $X_{t+1}$ follows a Bernoulli distribution conditioned on the value of $X_{t}$. Specifically we have
\[
    Pr\paren{X_{t+1} = 0 | X_t = 0} = .6, \quad Pr\paren{X_{t+1} = 0 | X_t = 1} = .4.
\]
Together with the specification $Pr\paren{X_1 = 0} = .5$, we can check that the process is stationary and ergodic. The reason that we choose this process as our first example is because it has a small set of causal states of size $3$. We list the causal states of sequences up to length $3$ in Tab.~\ref{tab:M2(.6,.4)CausalTable}.

\begin{table}[t]
    \centering
    \caption{Causality table of an order-$1$ Markov process with causal states $Q=\set{q_\lambda, q_0, q_1}$}
    \begin{tabular}{r|c|c|c}
        \hline
        $x$ & $Pr\paren{x}$ & $\phi_x$ & causal state \\ \hline
        \rowcolor{gray!50}$\lambda$ & $1$ & $(.5, .5)$ & $q_\lambda$\\
        $0$ & $.5$ & $(.6, .4)$ & $q_0$\\
        \rowcolor{MidLavender}$1$ & $.5$ & $(.4, .6)$ & $q_1$\\
        $00$ & $.3$ & $(.6, .4)$ & $q_0$\\
        \rowcolor{MidLavender}$01$ & $.2$ & $(.4, .6)$ & $q_1$\\
        $10$ & $.2$& $(.6, .4)$ & $q_0$\\
        \rowcolor{MidLavender}$11$ & $.3$& $(.4, .6)$ & $q_1$\\
        $000$ & $.18$ & $(.6, .4)$ & $q_0$\\
        \rowcolor{MidLavender}$001$ & $.12$ & $(.4, .6)$ & $q_1$\\
        $010$ & $.08$ & $(.6, .4)$ & $q_0$\\
        \rowcolor{MidLavender}$011$ & $.12$ & $(.4, .6)$ & $q_1$\\
        $100$ & $.12$ & $(.6, .4)$ & $q_0$\\
        \rowcolor{MidLavender}$101$ & $.08$ & $(.4, .6)$ & $q_1$\\
        $110$ & $.12$ & $(.6, .4)$ & $q_0$\\
        \rowcolor{MidLavender}$111$ & $.18$ & $(.4, .6)$ & $q_1$\\ 
        $\vdots$ & $\vdots$ & $\vdots$ & $\vdots$\\ \hline 
    \end{tabular}
    \vspace{.2cm}
    \label{tab:M2(.6,.4)CausalTable}
\end{table}
Since $\mu_{x}$ is defined on an infinite dimensional space, we only show the symbolic derivative $\phi_x$ in Tab.~\ref{tab:M2(.6,.4)CausalTable}, but we can verify that $\mu_x = \mu_y$ if and only if $\phi_x=\phi_y$ for this process.

\end{exmpl}

Now, we conceptualize the labeled directed graph obtained from analyzing the causal states by an automaton structure~\cite{yan1998introductionMachineComputation}, which we call probabilistic finite-state automaton~\cite{chattopadhyay2014data}, and show how we can get a stochastic process from it.
\begin{defn}[Probabilistic Finite-State Automaton (PFSA)]
A \textbf{probabilistic finite-state automaton} $G$ is specified by a quadruple $\paren{\Sigma, Q, \delta, \pitilde}$, where $\Sigma$ is a finite alphabet, $Q$ is a finite set of states, $\delta$ is a partial map from $Q\times\Sigma$ to $Q$ called transition map, and $\pitilde$, called observation probability, is a map from $Q$ to $\mathbf{P}_\Sigma$, where $\mathbf{P}_\Sigma$ is the space of probability distributions over $\Sigma$. The entry indexed by $\sigma$ of $\pitilde(q)$ is written as $\pitilde(q, \sigma)$.

\begin{figure}[t]
    \centering
    \includegraphics[scale=1]{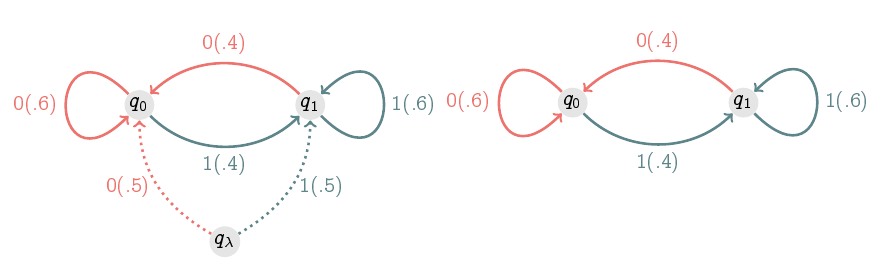}
    \caption{The graph on left is the labeled directed graph constructed from Tab.~\ref{tab:M2(.6,.4)CausalTable}. We note that the graph is \emph{not} strongly connected since $q_\lambda$ does not have any incoming edge. The graph on right is the strongly connected component of the graph on left.}
    \label{fig:M2_with_q_lambda_and_M2}
\end{figure}

We call the directed graph (not necessarily simple with possible loops and multiedges) with vertices in $Q$ and edges specified by $\delta$ the \textbf{graph of the PFSA} and, unless stated otherwise, we assume it to be \textbf{strongly connected}~\cite{bondy2008graph}, which means for any pair $q, q'\in Q$, there is a sequence $\sigma_1\sigma_2\cdots\sigma_k$, such that $\delta\paren{q_{i-1}, \sigma_i} = q_i$ for $i = 1, 2, \dots, k$ with $q_0=q$ and $q_k = q'$. 

To generate a sequence of symbols, assuming $G$'s current state is $q$, it then outputs symbol $\sigma$ with probability $\pitilde\paren{q, \sigma}$, and moves to state $\delta(q, \sigma)$. We see that $\delta$ is partial because $\delta(q, \sigma)$ is undefined when $\pitilde(q, \sigma) = 0$.
\end{defn}

\begin{defn}[Observation and Transition Matrices]
Given a PFSA $\paren{\Sigma, Q, \delta, \pitilde}$, the \textbf{observation matrix} $\Pitilde$ is the $\abs{Q}\times\abs{\Sigma}$ matrix with the $(q,\sigma)$-entry given by $\pitilde(q, \sigma)$, and the \textbf{transition matrix} $\Pi$ is the $\abs{Q}\times\abs{Q}$ matrix with the $(q, q')$-entry, written as $\pi(q, q')$, given by 
\[
    \pi(q, q') = \sum_{\set{\sigma: \delta(q, \sigma)=q'}}\pitilde(q, \sigma).
\]
\end{defn}
It is straightforward to verify that both $\Pi$ and $\Pitilde$ are stochastic, \ie nonnegative with rows of sum $1$.

\begin{rem}
We borrow the terms \emph{observation matrix} and \emph{transition matrix} from the study of HMM~\cite{stamp2004revealing}. However, we need to point out here that our model differs from the HMM in that, in HMM, the transition from the current state to the next one is independent of the symbol generated by the current state, while in PFSA, the current state and symbol generated together determine the next state the PFSA will be in. 
\end{rem}

Unless specified otherwise, we assume the initial distribution to be the \textbf{stationary distribution}~\cite{kai1967markov_StDis} of $\Pi$. We denote the stationary distribution of $G$ by $\mathbf{p}_{G}$, or by $\mathbf{p}$ if $G$ is understood.

\begin{thm}
\label{thm:PFSAGeneratesStationaryErgodicProc}
Stochastic process generated by a PFSA $G$ with distribution on states initialized with $\mathbf{p}_{G}$ is stationary and ergodic.
\end{thm}
\vspace{-.2cm}
\textit{proof omitted.}
\vspace{.1cm}

Example \ref{exmpl:M2} shows that we may derive a PFSA from a stationary ergodic process, and Thm.~\ref{thm:PFSAGeneratesStationaryErgodicProc} shows that the process generated by the PFSA thus obtained is also stationary and ergodic. This motivates us to seek a characterization for stochastic processes that gives rise to a PFSA. Since the process in Example \ref{exmpl:M2} is an order-$1$ Markov process, which is the simplest non-i.i.d.~process, it is legitimate to ask whether a process has to be Markov to have a PFSA generator. This desired characterization is obtained from studying the properties of causal states, which we do in the next section.

\begin{rem}
Table~\ref{tab:CM_PFSA_HMM} compares the three generative models of stochastic processes mentioned in this paper: Markov chain(MC), PFSA, and hidden Markov model(HMM). We note that a Markov chain produces a sequence of states, while sequences produced by PFSA and hidden Markov model take values in their respect output alphabets. We can also see that HMM can be considered as an extension to MC by adding an output alphabet and observation probabilities while PFSA are not directly comparable to either MC or HMM.
\begin{table}[t]
    \centering
   \caption{Comparing Markov chain, PFSA, and hidden Markov model.}
    \label{tab:CM_PFSA_HMM}
    \begin{tabular}{p{.033\textwidth}|p{.18\textwidth}|p{0.195\textwidth}}
        \hline
        Model & Defining variables & Example\\ \hline \hline
        MC &
        \begin{minipage}[h]{.18\textwidth}
            Set of states;\\
            Transition probabilities.
        \end{minipage} & 
        \begin{minipage}[h]{0.195\textwidth}
            \centering
            \includegraphics[scale=1]{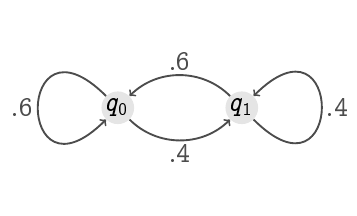}
        \end{minipage} 
        \\ \hline
        PFSA &
        \begin{minipage}[h]{.18\textwidth}
            Set of states;\\
            Output alphabet;\\
            Transition function;\\
            Observation probabilities.
        \end{minipage} & 
        \begin{minipage}[h]{0.195\textwidth}
            \centering
            \vspace{.2cm}
            \hspace*{-4pt}
            \includegraphics[scale=.9]{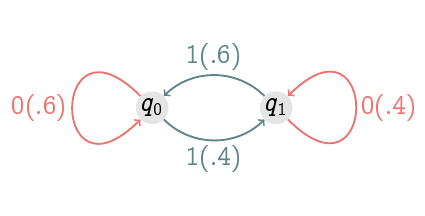}
        \end{minipage}
        \\ \hline
        HMM &
        \begin{minipage}[h]{.18\textwidth}
            Set of states;\\
            Output alphabet;\\
            Transition probabilities;\\
            Observation probabilities.
        \end{minipage}& 
        \begin{minipage}[h]{0.195\textwidth}
            \centering
            \vspace{-.2cm}
            \includegraphics[scale=1.]{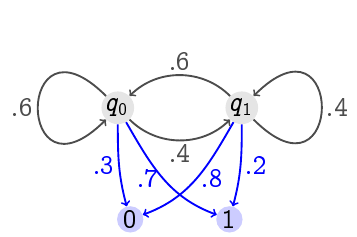}
        \end{minipage}
        \\ \hline 
    \end{tabular}
 \end{table}
\end{rem}

\section{Stochastic Processes with PFSA Generator}
\label{sec:PFSAGenerator}
\subsection{Persistent Causal States}
\label{subsec:PersistentCausalStates}
\begin{defn}[Persistent and Transient Causal States]
Let $Q$ be the set of causal states of a stationary ergodic process. For every $q\in Q$ and $d\in\mathbb{N}$, let $p_{d}(q) = \phi^{d}\set{[x] = q}$, \ie the probability of length-$d$ sequences who are equivalent to $q$. A causal state $q$ is \textbf{persistent} if $\liminf_{d\rightarrow\infty}p_{d}(q) >0$, and \textbf{transient} if otherwise. We denote the set of persistent causal states by $Q^{+}$.
\end{defn}

\begin{rem}
Here we borrow the term \emph{transient state} from Markov chains literature, for example \cite{gagniuc2017markov}, but we should note that the two concepts are not identical. A Markov chain never revisits transient states as soon it hits a recurrent state. However, although a transient causal state could never be revisited, as the $q_\lambda$ in Example \ref{exmpl:M2}, it could also be revisited for infinitely many times. The transient states in MC and PFSA are similar in that the probability of a Markov chain being in a transient state diminishes as time increases, and a transient causal states also has $\liminf_{d\rightarrow\infty}p_{d}(q) = 0$. Since transient states can recur, we name the counterpart to transient causal state in PFSA by \emph{persistent causal state}, not by recurrent state as in MC.
\end{rem}

For any pair $q, q'\in Q$, let $\pi_{q, q'} = \sum_{\set{\sigma: q\sigma = q'}}\phi_{q}(\sigma)$, where the expression $q\sigma = q'$ is a shorthand for $[x\sigma] = q'$ for all $[x] = q$. The following proposition shows that $\pi_{q, q'}$ captures the flow of probability over causal states as sequence length increases.  
\begin{prop}
We have $p_{d}(q') = \sum_{q\in Q}\pi_{q, q'}p_{d-1}(q)$ for each $q'\in Q$ and $d \in \mathbb{N}^{+}$. Furthermore, there is no flow from a persistent state to a transient one, \ie $\pi_{q, q'} = 0$ for $q\in Q^{+}$ and $q'\in Q\setminus Q^{+}$.
\end{prop}
\vspace{-.2cm}
\textit{proof omitted.}
\vspace{.1cm}

\begin{thm}
\label{thm:PFSAOnStateSetQPlus}
Let $Q^{+}$ be the set of persistent causal states of a stationary ergodic process $\mathscr{P}$. Then, $p(q) = \lim_{d\rightarrow\infty}p_{d}(q)$ exists for every $q\in Q^{+}$. Furthermore,  if $Q^{+}$ is finite and $\sum_{q\in Q^{+}} p(q) = 1$, the process generated by the PFSA $G = \paren{\Sigma, Q^{+}, \delta, \pitilde}$ with $\delta(q, \sigma) = q\sigma$ and $\pitilde(q, \sigma) = \phi_{q}(\sigma)$ is exactly $\mathscr{P}$. In fact, we have $\left.\mathbf{p}_{G}\right|_q = p(q)$ for $q\in Q^{+}$.
\end{thm}
\vspace{-.2cm}
\textit{proof omitted.}
\vspace{.1cm}

\begin{exmpl}[An Order-Two Markov Process]
\label{exmpl:M4}
Now, let us consider an order-$2$ Markov process over alphabet $\Sigma = \set{0, 1}$, in which $X_{t+2}$ follows a Bernoulli distribution conditioned on the value of $X_tX_{t+1}$. More specifically, denoting $Pr\paren{X_{t+2} = 0 | X_{t}X_{t+1} = ij}$ by $p_{ij}$ for $i, j\in\set{0,1}$, we have $p_{00} = .3$, $p_{01} = .2$, $p_{10}=.8$, $p_{11} = .7$. Together with the specification $\Pr\paren{X_2 = 0|X_1 = 0} = 8/15$, $Pr\paren{X_2 = 0|X_1 = 1} = 7/15$, and $Pr\paren{X_1 = 0} = .5$, we can check that the process is stationary and ergodic. We list the causal states of sequences up to length $3$ in Tab.~\ref{tab:M4(.3,.2,.8,.7)CausalTable}.
\begin{table}[t]
    \centering
    \caption{Causality table of an order-$2$ Markov process with causal states $Q=\set{q_\lambda, q_0, q_1, q_{00}, q_{01}, q_{10}, q_{11}}$. We can see that the causal states $q_{00}$, $q_{01}$, $q_{10}$, and $q_{11}$ are named after the last two symbols of the corresponding sequences, which is a demonstration of the order-$2$ Markovity of the process, \ie the distribution of future events is determined completely by length-$2$ immediate history.}
    \begin{tabular}{r|c|c|c}
        \hline
        $x$ & $Pr(x)$ & $\phi_x$ & causal state \\ \hline
        \rowcolor{gray!20}$\lambda$ & $1$ & $(1/2, 1/2)$ & $q_{\lambda}$ \\
        \rowcolor{gray!40}$0$ & $1/2$ & $(8/15, 7/15)$ & $q_{0}$ \\
        \rowcolor{gray!60}$1$ & $1/2$ & $(7/15, 8/15)$ & $q_{1}$ \\
        \rowcolor{MidLavender!20}$00$ & $4/15$ & $(3/10, 7/10)$ & $q_{00}$ \\
        \rowcolor{MidLavender!40}$01$ & $7/30$ & $(1/5, 4/5)$ & $q_{01}$ \\
        \rowcolor{MidLavender!80}$10$ & $7/30$ & $(4/5, 1/5)$ & $q_{10}$ \\
        \rowcolor{MidLavender}$11$ & $4/15$ & $(7/10, 3/10)$ & $q_{11}$ \\
        \rowcolor{MidLavender!20}$000$ & $2/25$ & $(3/10, 7/10)$ & $q_{00}$ \\
        \rowcolor{MidLavender!40}$001$ & $14/75$ & $(1/5, 4/5)$ & $q_{01}$ \\
        \rowcolor{MidLavender!80}$010$ & $7/150$ & $(4/5, 1/5)$ & $q_{10}$ \\
        \rowcolor{MidLavender}$011$ & $14/75$ & $(7/10, 3/10)$ & $q_{11}$ \\
        \rowcolor{MidLavender!20}$100$ & $14/75$ & $3/10, 7/10$ & $q_{00}$ \\
        \rowcolor{MidLavender!40}$101$ & $7/150$ & $(1/5, 4/5)$ & $q_{01}$ \\
        \rowcolor{MidLavender!80}$110$ & $14/75$ & $(4/5, 1/5)$ & $q_{10}$ \\
        \rowcolor{MidLavender}$111$ & $2/25$ & $(7/10, 3/10)$ & $q_{11}$ \\
        $\vdots$ & $\vdots$ & $\vdots$ & $\vdots$\\ \hline 
    \end{tabular}
    \vspace{.2cm}
    \label{tab:M4(.3,.2,.8,.7)CausalTable}
\end{table}

Since $\mu_{x}$ is defined on an infinite dimensional space, we only show $\phi_x$ in Tab.~\ref{tab:M4(.3,.2,.8,.7)CausalTable}, but we can check that $\phi_x = \phi_y$ if and only if $\mu_{x} = \mu_{y}$ for this process. Since $q_\lambda$, $q_0$, $q_1$ only show up once, while $q_{00}$, $q_{01}$, $q_{10}$, $q_{11}$ appear repeatedly, we have $Q^{+} = \set{q_{00}, q_{01}, q_{10}, q_{11}}$. With more detailed calculation, we can show that $p\paren{q_{00}} = 4/15$, $p\paren{q_{01}} = 7/30$, $p\paren{q_{10}} = 7/30$, and $p\paren{q_{11}} = 4/15$, which sum up to $1$. According to Thm.~\ref{thm:PFSAOnStateSetQPlus}, we can construct a PFSA with state set $Q^{+}$ that generates exactly the same process. We demonstrate the labeled directed graph constructed on $Q$ in Fig.~\ref{fig:M4}, and the PFSA is exactly the induced subgraph \cite{ray2012graph} on $Q^{+}$, which is also the unique strongly connected component of the graph. We can show that the stationary distribution of the PFSA is exactly $(4/15, 7/30, 7/30, 4/15)$. 

\begin{figure}[t]
    \centering
    \includegraphics[scale=1.]{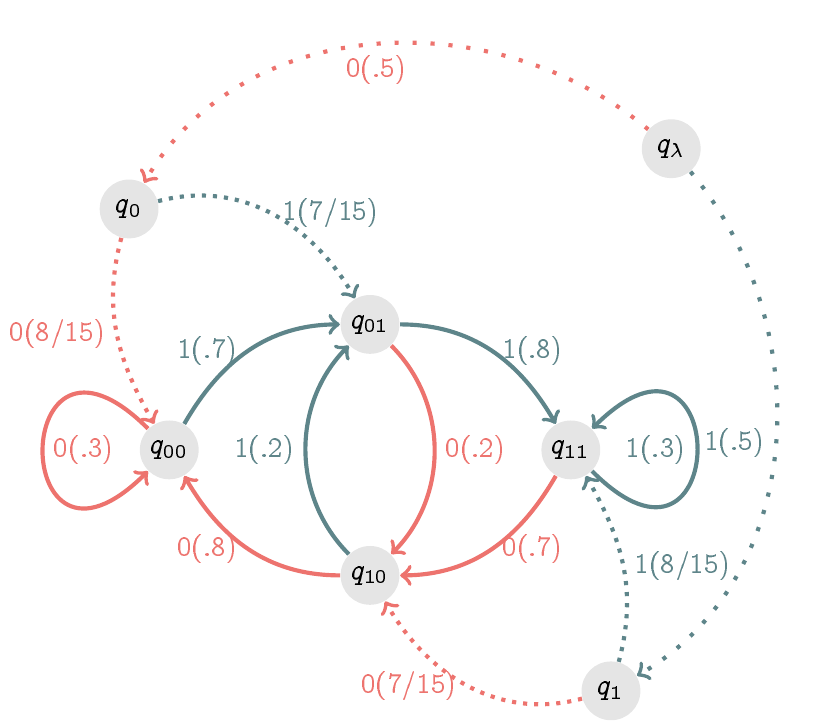}
    \caption{Labeled directed graph obtained from Tab.~\ref{tab:M4(.3,.2,.8,.7)CausalTable}. The edges from transient causal states are dotted, while those from persistent states are solid.}
    \label{fig:M4}
\end{figure}
\end{exmpl}

\begin{figure}[t]
    \centering
    \includegraphics[scale=1.]{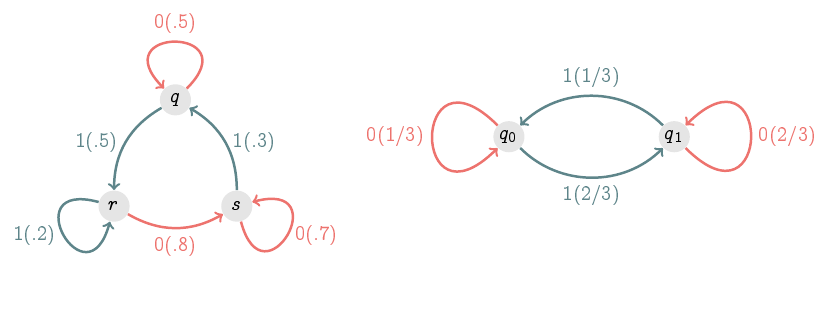}
    \caption{The PFSA $T$ on left generates a stochastic process with infinite $Q$ while $Q^{+}$ of size $3$. The PFSA $S$ on right generates a stochastic process with empty $Q^{+}$ but $\overline{Q}^{+}$ (defined in Sec.~\ref{subsec:AccumulationCausalStates}) of size $2$.}
    \label{fig:T_and_S2}
\end{figure}

\begin{figure}[t]
    \centering
    \begin{tikzpicture}[scale=.43, every node/.style={transform shape}]
        \node[
            label={[rotate=90, xshift=1.9cm]left:{\Huge frequency}},
            label={[xshift=.5cm, yshift=.3cm]below:{\Huge sequence length}}
        ] at (-5, 0) {\includegraphics[scale=.95]{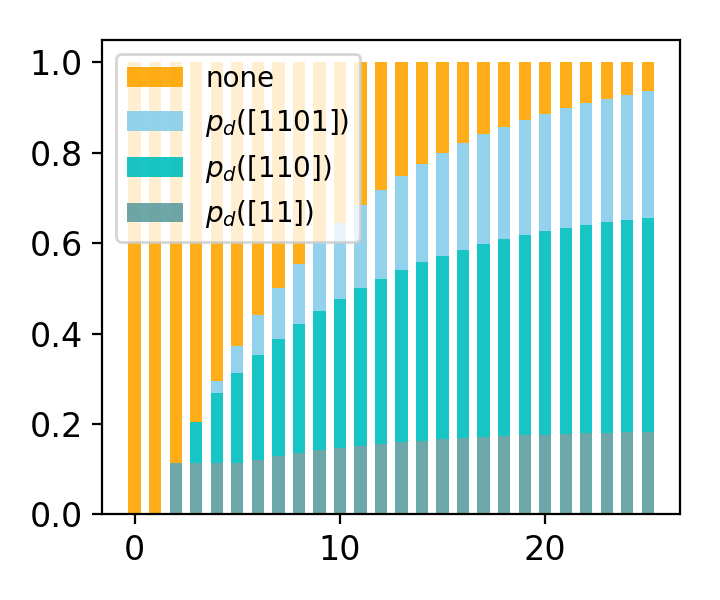}};
        \node[
            label={[rotate=90, xshift=1.5cm, yshift=.2cm]left:{\Huge frequency}},
            label={[xshift=.5cm, yshift=.3cm]below:{\Huge $n$ as in Eq.~(2)}} 
        ] at (5, .3) {\includegraphics[scale=1., trim=0 0 25 0, clip]{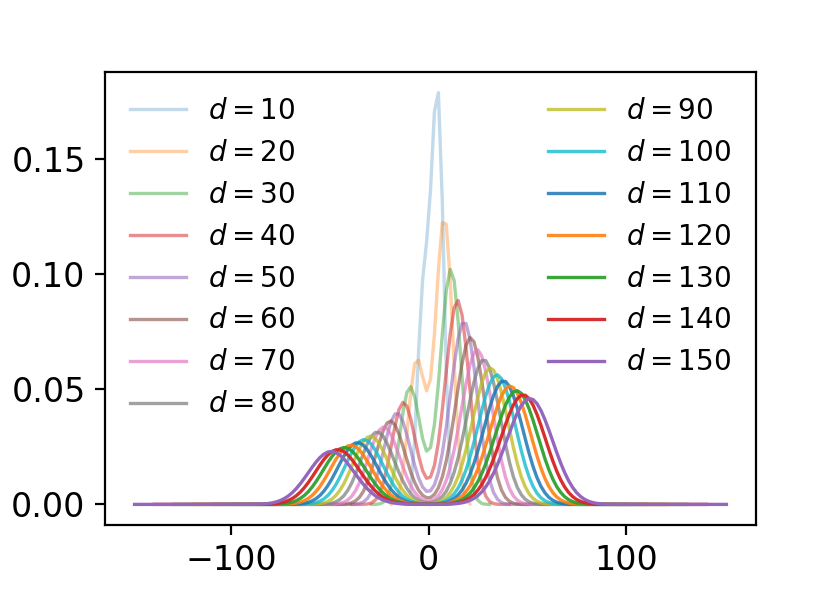}};
    \end{tikzpicture}
    \caption{The bar plot on left shows the probabilities of $[11]$, $[110]$, and $[1101]$ for sequence length $d=0, 1, \dots, 25$ as discussed in Example \ref{exmpl:3StateSyncPFSA}. The curves on right show the contour of $p_d\paren{q_n}$ against $n$ for sequence length $d = 10, 20, \dots, 150$ as discussed in Example \ref{exmpl:S2}.}
    \label{fig:T_QPlus_S2_Accum}
\end{figure}

\begin{exmpl}[A PFSA on Three States]
\label{exmpl:3StateSyncPFSA}
In this example, we analyze the stochastic process generated by the PFSA on the right of Fig.~\ref{fig:T_and_S2}. We nickname the PFSA by $T$. We show that $Q$ of this process is infinite, while $Q^{+}$ is of size $3$. We first notice that, no matter what state the PFSA resides, the sequence $11$, and hence any sequence ending in $11$, will take it to state $r$, which generates symbol $0$ with probability $.8$, and $1$ with probability $.2$. We also note that, whenever there are two consecutive $1$s in a given sequence in $\Sigma^{\star}$, we know for sure the state the PFSA resides. For example, sequence $110$ will take the PFSA to $s$, and $1101$, to $q$. On the left of Fig.~\ref{fig:T_QPlus_S2_Accum}, we show the probabilities of causal states $[11]$, $[110]$, $[1101]$, and the sum of probabilities of all other causal states for sequence length $d = 0, \dots, 25$. We see from the bar plots that the sum of concentrations of $[11]$, $[110]$, and $[1101]$ approaches $1$ as $d$ increases. We also point out that, with all numbers rounded up to three decimal places, $p_{25}\paren{[11]}=0.182$, $p_{25}\paren{[110]} = 0.474$, and $p_{25}\paren{[1101]} = 0.279$, while the stationary distributions of the states $r$, $s$, and $q$ are $0.190, 0.506, 0.304$, respectively.

However, we also note that $Q$ of the process is actually infinite by observing the fact that $\phi_{\bracket{0^{d}}}$, where $\sigma^{d}$ means $\sigma$ repeated $d$ times, are all distinctive.

We note that the process generated by this PFSA is \emph{not} Markov, as implied by the infinity of $Q$. However, the fact that there are only three persistent causal states whose sum of probabilities approaches $1$  allows it to have  a PFSA generator.    
\end{exmpl}

\begin{exmpl}[A Stochastic Process with Empty $Q^{+}$]
\label{exmpl:S2}
In this example, we analyze the stochastic process generated by the PFSA on the left of Fig.~\ref{fig:T_and_S2}. We nickname the PFSA by $S$. We show that $Q$ of this process is infinite while $Q^{+}$ is empty. Without run into details of the computation, we point out the fact that causal states of this process are also uniquely characterized by their symbolic derivatives, and the set $\set{\phi(q)|q\in Q}$ is in one-to-one correspondence with $\mathbb{Z}$. More specifically, we have
\begin{equation}
\label{eq:CausalStatesOfS2}
    Q = \set{q_n\left|\phi_{q_n}\propto\paren{.5^{n + 1} + 1, .5^{n} + .5}, n\in\mathbb{Z}\right.},
\end{equation}
where $\propto$ means being proportional to, and 
\[
    \pi_{q_{n}, q_{n+1}} = \frac{2}{3}\frac{.5^{n+1} + 1}{.5^{n} + 1},\quad\pi_{q_{n}, q_{-n+1}} = \frac{1}{3}\frac{.5^{n-1} + 1}{.5^{n} + 1},
\]
with $\pi_{q_{n}, q_{n+1}} + \pi_{q_{n}, q_{-n+1}} = 1$, for all $n\in\mathbb{Z}$. We demonstrate on the left of Fig.~\ref{fig:T_QPlus_S2_Accum} the contour of $p_{d}\paren{q_n}$ against $n$ for sequence length $d = 10, 20, \dots, 150$. It takes some more work to show rigorously, but we can speculate that, for any fixed $n\in\mathbb{Z}$, $p_{d}\paren{q_n}$ approaches $0$ as $d$ approaches infinity, as the curves flatten out with increasing $d$.
\end{exmpl}

\subsection{Accumulation Causal States}
\label{subsec:AccumulationCausalStates}
We see from Example \ref{exmpl:S2} that we can have a PFSA that generates stochastic process with empty $Q^{+}$. In such a case, can we still get the PFSA structure back by studying the the set of causal states of the process? The answer is yes. 

\begin{defn}[Epsilon-Ball of Measure]
Denote the collection all measures on $\paren{\Sigma^{\omega}, \F}$ by $\mathcal{M}_{\Sigma}$, and let $\nu\in\mathcal{M}_{\Sigma}$, the $\varepsilon$-ball of order $d$ centered at $\nu$ is defined by
\[
    B_{d, \varepsilon}(\nu) = \set{\nu'\in\mathcal{M}_{\Sigma}\left| \sum_{x\in\Sigma^{d}}\abs{\nu'\paren{x\Sigma^{\omega}} - \nu\paren{x\Sigma^{\omega}}} < \varepsilon \right.}.
\]
In another words, $B_{d, \varepsilon}$ is the collection of all measures that is no more than $\varepsilon$ away from $\nu$ with respect to total variation distance over $\Sigma^{d}$.
\end{defn}

\begin{defn}[Accumulation Causal States]
Let $Q$ be the set of causal states of a stochastic process $\mathscr{P}$, a measure $\nu\in\mathcal{M}_{\Sigma}$ is an \textbf{accumulation causal state} of $\mathscr{P}$ if 
\[
    p_{l, d, \varepsilon}(\nu) = \phi^{l}\set{\mu_{[x]}\in B_{d, \varepsilon}(\nu)}
\]
satisfies $p_{d,\varepsilon}(\nu) = \liminf_{l\rightarrow\infty}p_{l, d, \varepsilon}(\nu) > 0$ for all $d\in\mathbb{N}^{+}$ and $\varepsilon > 0$. That is, a measure $\nu$ is an accumulation causal state if, no matter how large $d$ is and how small $\varepsilon$ is, the sum of probabilities of length-$l$ sequences falling in $B_{d, \varepsilon}(\nu)$ does not vanish as $l$ approaches infinity.

The collection of accumulation causal states is denoted by $\overline{Q}$. Since $p_{d, \varepsilon}(\nu)$ is monotonically decreasing as $d \rightarrow \infty$ and $\varepsilon \rightarrow 0$, $p(\nu)=\lim_{d\rightarrow\infty}\lim_{\varepsilon\rightarrow0}p_{d, \varepsilon}(\nu)$ is well-defined. A measure $\nu$ with $p(\nu) > 0$ is called an \textbf{atomic accumulation causal state}, and the collection of all atomic accumulation causal states is denoted by $\overline{Q}^{+}$. 
\end{defn}

\begin{defn}[Translation Measure]
Let $\nu\in\mathcal{M}_{\Sigma}$, the translation of $\nu$ by $\sigma$ for $\nu\paren{\sigma\Sigma^{\omega}} > 0$, denoted by $\nu_{\sigma}$, is the extension to $\F$ of the premeasure on the semiring $\SR$ given by
\[
    \nu_{\sigma}\paren{x\Sigma^{\omega}} = \frac{\nu\paren{\sigma x\Sigma^{\omega}}}{\nu\paren{\sigma\Sigma^{\omega}}}.
\]
\end{defn}

\begin{prop}
$\overline{Q}^{+}$ is closed under translation.
\end{prop}
\vspace{-.2cm}
\textit{proof omitted.}
\vspace{.1cm}

\begin{thm}
Let $\mathscr{P}$ be a stationary ergodic stochastic process with finite $\overline{Q}^{+}$ and $\sum_{\nu\in \overline{Q}^{+}}p(\nu) = 1$. Then the process generated by the PFSA $G = \paren{\Sigma, \overline{Q}^{+}, \delta, \pitilde}$ with $\delta(\nu, \sigma) = \nu_{\sigma}$ and $\pitilde\paren{\nu, \sigma} = \nu\paren{\sigma\Sigma^{\omega}}$ is exactly $\mathscr{P}$. In fact, we have $\left.\mathbf{p}_{G}\right|_{\nu} = p(\nu)$. 
\end{thm}
\vspace{-.2cm}
\textit{proof omitted.}
\vspace{.1cm}

\begin{exmpl}[Example \ref{exmpl:S2} Revisited]
\label{exmpl:S2Revisit}
We demonstrate that $\overline{Q}^{+}$ of the process in Example \ref{exmpl:S2} has two elements, again by observation. We plot the cumulative probability density functions of $\phi_{q}(0)$ for each sequence length $d = 10, 20, 30, 40$ in Fig.~\ref{fig:S2AccumulativeProbsBy10}. More specifically, for each fixed $d$, the $x$-coordinates of the dots are in $\Phi^{d} =\set{\phi_{[x]}(0)\left|x\in\Sigma^{d}\right.}$, while the $y$-coordinate of a dot with $x$-coordinate $h_0\in \Phi^{d}$ equals $\phi^{d}\paren{\set{\phi_{[x]}(0) \leq h_0}}$. We can see clearly that, the cumulative function converges to a step function with steps at $1/3$ and $2/3$ as $d$ increases. The fact implies that $\nu\in\overline{Q}^{+}$ satisfies that $\nu\paren{0\Sigma^{\omega}}$ is either $1/3$ or $2/3$. We see from \eqref{eq:CausalStatesOfS2} that the two measures in $\overline{Q}^{+}$ are exactly $q_{-\infty}$ and $q_{\infty}$. Fig.~\ref{fig:S2AccumulativeProbsBy10} also implies that that $p\paren{q_{-\infty}} = 1/3$ and $p\paren{q_{\infty}} = 2/3$, which is exactly the stationary distribution on the state set of $S$.
\begin{figure}[t]
    \centering
    \begin{tikzpicture}[scale=.305, every node/.style={transform shape}]
        \node[
            label={[align=center, rotate=90, yshift=.5cm, xshift=1.5cm]left:{\Huge cumulative\\\Huge density}},
            label={[yshift=.2cm, xshift=0cm]below:{\Huge $\phi_q(0)$}}
        ] at (0, 0) {\includegraphics[scale=1]{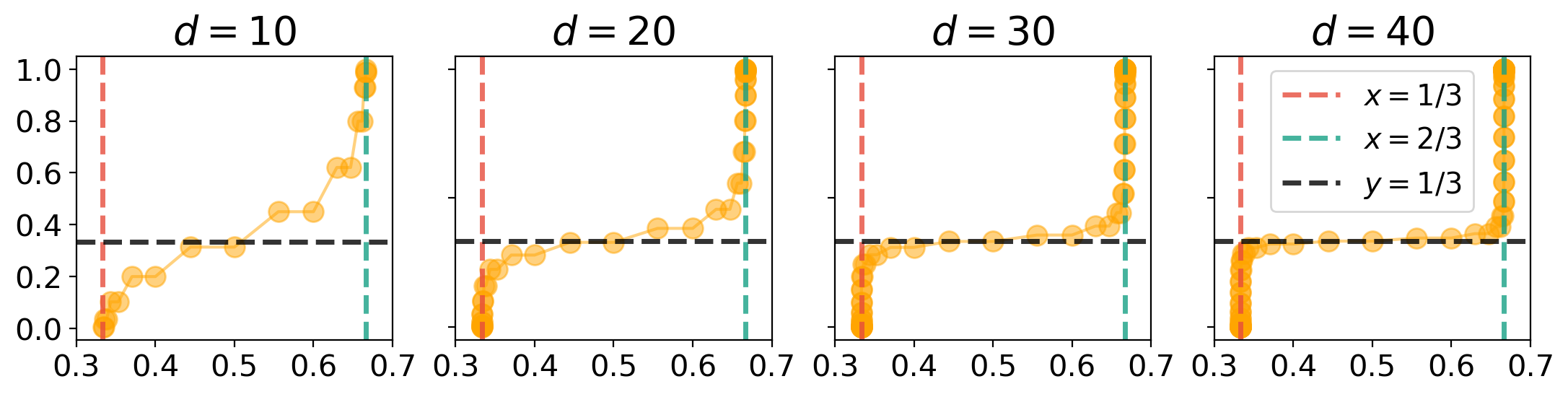}};
    \end{tikzpicture}
    \caption{Cumulative probability density function of $\phi_{q}(0)$.}
    \label{fig:S2AccumulativeProbsBy10}
\end{figure}
\end{exmpl}

\section{Inference algorithm of PFSA}
\label{sec:inference_PFSA}
From the discussion in Sec.~\ref{sec:PFSAGenerator}, we see that a stochastic process has a PFSA generator if either finitely many causal states get all the probability in the limit, as described in Sec.~\ref{subsec:PersistentCausalStates}, or there exist finitely many measures in $\mathcal{M}_\Sigma$ whose arbitrarily small neighborhoods are populated by almost all the causal states in the limit, as described in Sec.~\ref{subsec:AccumulationCausalStates}. The implication of these observations goes beyond the theory of PFSA, and guide us through the designing of inference algorithms of the model. In fact, a valid heuristic of the inference algorithm of PFSA would be to apply any clustering algorithm to the set of causal states corresponding to sequences up to a certain length, and use the center of the clusters to serve as estimates to the states. However, this primitive heuristic has a drawback since the cluster structure of $\set{[x]|x\in\Sigma^{d}}$ may not be clear enough to facilitate a clustering algorithm. In order to get better estimates of the states, we need to fine tune our view into the set of causal states using the notion of $\varepsilon$-synchronizing sequence~\cite{chattopadhyay2014causality}.

\subsection{Epsilon-synchronizing Sequences}
Before introducing $\varepsilon$-synchronizing sequence, we first introduce the concept of observation induced distributions over the state set. Let $G$ be a PFSA, we know that the initial distribution over states is exactly the stationary distribution $\mathbf{p}_{G}$. Let us assume that the first symbol generated by $G$ is $\sigma$, denote by $\mathbf{p}_{G}(\sigma)$ the distribution over states after $G$ producing $\sigma$, we have
\[
    \left.\mathbf{p}_{G}(\sigma)\right|_{q} = \frac{1}{Z}\sum_{\set{q'|\delta\paren{q',\sigma} = q}}\pitilde(q', \sigma)\mathbf{p}_{G}|_{q'},
\]
where 
\[
    Z = \sum_{q \in Q}\sum_{\set{q'|\delta\paren{q',\sigma} = q}}\pitilde(q', \sigma)\mathbf{p}_{G}|_{q'},
\]
is the normalizer. 

\begin{defn}[Observation induced distributions]
Let $x=\sigma_{1}\dots\sigma_{n}$ be a sequence observed, the distribution over states induced by $x$ is defined inductively by
\[
    \left.\mathbf{p}_{G}\paren{\sigma_1\dots\sigma_{i}}\right|_{q} = \frac{1}{Z}\sum_{\set{q'|\delta\paren{q',\sigma_i} = q}}\pitilde(q', \sigma)\mathbf{p}_{G}\paren{\sigma_1\dots\sigma_{i-1}}|_{q'},
\]
where 
\[
    Z = \sum_{q \in Q}\sum_{\set{q'|\delta\paren{q',\sigma_i} = q}}\pitilde(q', \sigma)\mathbf{p}_{G}\paren{\sigma_1\dots\sigma_{i-1}}|_{q'},
\]
for $i = 1, \dots, n$, with the base case $\mathbf{p}_{G}(\lambda) = \mathbf{p}_{G}$.
\end{defn}

\begin{defn}[$\varepsilon$-synchronizing sequence]
Let $G$ be a strongly connected PFSA on state set $Q$ over alphabet $\Sigma$. A sequence $x\in\Sigma^{\star}$ is called an \textbf{$\varepsilon$-synchronizing sequence} for some $\varepsilon>0$ if there exists a $q \in Q$ such that $\norm{\mathbf{p}_{G}(x) - \mathbf{e}_q}_\infty <\varepsilon$, where $\mathbf{e}_q$ is the base probability vector with the entry indexed by $q$ equalling $1$. 
\end{defn}

The reason that the $\varepsilon$-synchronizing sequences are important to inference is that $\set{[x_{\varepsilon}x]| x\in\Sigma^{d}}$ tends to have a much clearer cluster structure than $\set{[x]|x\in\Sigma^{d}}$ for an $\varepsilon$-synchronizing sequence $x_{\varepsilon}$. 

\subsection{\algo~Algorithm}
We give a brief review to the algorithm called \algo~proposed in~\cite{chattopadhyay2013abductive} in this section. By a sub-sequence, we mean a \emph{consecutive} sub-sequence.
\begin{defn}[Empirical Symbolic Derivatives]
Let $x\in\Sigma^{\star}$, the empirical symbolic derivative $\hat{\phi}^{x}_y$ of a  sub-sequence $y$ of $x$ is given by 
\[
    \hat{\phi}^{x}_{y}(\sigma) = \frac{\textrm{number of sub-sequence $y\sigma$ in $x$}}{\textrm{number of sub-sequence $y$ in $x$}},
\]
for all $\sigma\in\Sigma$.
\end{defn}
Our inference algorithm is called \algo~for \underline{Gen}erator \underline{E}xtraction Using \underline{Se}lf-\underline{S}imilar \underline{S}emantics,  With the input of a long enough observed sequence $x$, \algo~takes the following three steps to infer a PFSA: 

\vspace{.6mm}
\noindent\textbf{Step one: Approximate $\varepsilon$-synchronizing sequence:} Calculate 
\[
    \mathcal{D}^{x}_{\varepsilon} = \set{\hat{\phi}^{x}_y \left| \textrm{$y$ is a sub-sequence of $x$ with } |y|\leq \log_{\abs{\Sigma}}\frac{1}{\varepsilon}\right.},
\]
Then, select a sequence $x_\varepsilon$ with $\hat{\phi}^{x}_{x_\varepsilon}$ being a vertex of the convex hull of $\mathcal{D}^{x}_{\varepsilon}$.

\vspace{.6mm}
\noindent\textbf{Step Two: Identify transition structure:} For each state $q$, we associate a sequence identifier $x_q \in x_\varepsilon\Sigma^\star$, and a probability distribution $d_q$ on $\Sigma$. We extend the structure recursively: Initialize the state set as $Q = \set{q_0}$, find $x_{q_0}$ and set $d_{q_0} = \hat{\phi}^x_{x_{q_0}}$; Calculate the empirical symbolic derivative of $x_q\sigma$ for each state $q \in Q$ and $\sigma\in\Sigma$. If $\norm{\hat{\phi}^x_{x_q\sigma} - d_{q'}}_\infty \leq \varepsilon$ for some $q' \in Q$, then define $\delta(q,\sigma) = q'$. However, if no such $q'$ exists in $Q$, add a new state $q'$ to $Q$, and define $x_{q'} = x_q\sigma$, and $d_{q'} = \hat{\phi}^x_{x_q\sigma}$. The process terminates when no more states can be added to $Q$. The inferred PFSA is the strongly connected component of the directed graph thus obtained.

\vspace{.6mm}
\noindent\textbf{Step Three: Identify observation probabilities:} Initialize counter $N^{q}_\sigma$ for each state $q$ and symbol $\sigma$; choose an arbitrary initial state in the graph obtained in step two and run sequence $x$ through it, \ie if current state is $q$, and the next symbol from $x$ is $\sigma$, then move to $\delta(q,\sigma)$, and add $1$ to counter $N^{q}_\sigma$; finally, calculate the observation probability map by $\pitilde(q) = \nrm{\paren{N^q_\sigma}_{\sigma\in\Sigma}}$.

\section{Entropy Rate and KL Divergence}
\label{sec:Entropies}
\subsection{Irreducibility of PFSA}
\label{subsec:Irreducibility}
We first discuss the concept of irreducibility for PFSA.
\begin{defn}
A PFSA $G$ is irreducible if there is no other PFSA with strictly fewer number of states that generates the same stochastic process as $G$ does.  
\end{defn}
The definition of PFSA itself doesn't ensure irreducibility, as shown by example~\ref{exmpl:NonReducedPFSA}.

\begin{exmpl}[Reducible PFSA]
\label{exmpl:NonReducedPFSA}
In Fig.~\ref{fig:reduciblePFSA}, we show two reducible PFSA. The PFSA on left generates the same process as the PFSA on the right Fig.~\ref{fig:M2_with_q_lambda_and_M2} does, while the PFSA on right generates the same procces as the PFSA on the right of Fig.~\ref{fig:T_and_S2} does, but both with one more state than their respective irreducible versions.
\begin{figure}[!t]
    \centering
    \begin{tikzpicture}[scale=.64, every node/.style={transform shape}]
        \node at (-3.4, 0) {\includegraphics[trim={0 0 0 10}, clip]{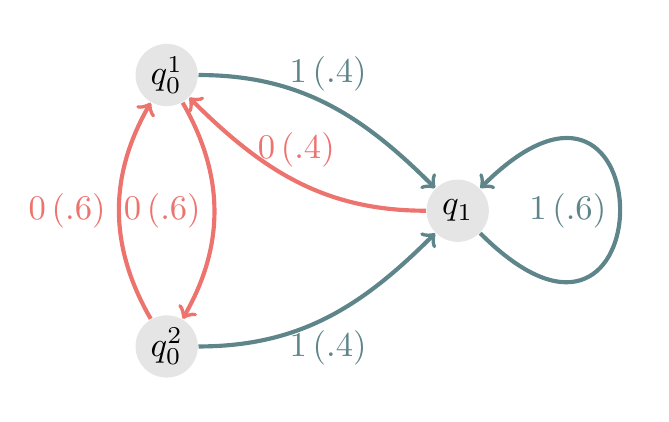}};
        \node at (3.4, 0) {\includegraphics[trim={0 0 0 10}, clip]{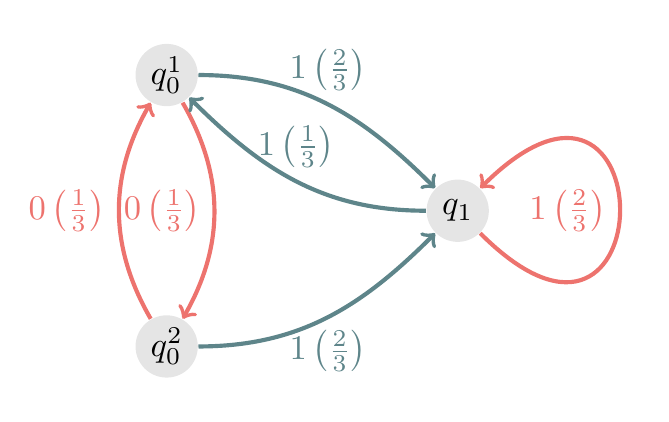}};
    \end{tikzpicture}
    \caption{Reducible PFSA.}
    \label{fig:reduciblePFSA}
\end{figure}
\end{exmpl}

\begin{defn}[Measure of state $\mu_{q}$ and Equivalent States]
Let a PFSA be specified by the quadruple $\paren{\Sigma, Q, \delta, \pitilde}$ and the measure $\mu_{q}$ be defined by $\mu_q\paren{x\Sigma^{\omega}} = \pitilde(q, x)$. Two states $q, q'\in Q$ are \textbf{equivalent} if and only $\mu_{q} = \mu_{q'}$. 

\end{defn}
We note that $q_{0}^{1}$ and ${q}_0^2$ in both PFSA in Fig.~\ref{fig:reduciblePFSA} are equivalent. We also see that we can get the corresponding irreducible PFSA back by collapsing equivalent states to a single state.

\begin{thm}[Characterization of Irreducibility]
A PFSA is irreducible if and only if it has no equivalent states. Furthermore, a irreducible PFSA is unique in the sense that, if two irreducible PFSA $G_1 = \paren{\Sigma, Q_1, \delta_1, \pitilde_1}$ and $G_2 = \paren{\Sigma, Q_2, \delta_2, \pitilde_2}$ generate the same stochastic process, there must be a one-to-one correspondence $f:Q_1\rightarrow Q_2$ such that $f\paren{\delta_1(q,\sigma)}=\delta_2(f(q), \sigma)$ and $\pitilde_2(f(q)) = \pitilde_1(q)$.
\end{thm}

\begin{cor}
The PFSA constructed on the set of persistent states $Q^{+}$ and the set of atomic accumulation states $\overline{Q}^{+}$ are irreducible.
\end{cor}

\subsection{Entropy Rate and KL Divergence}
\label{subsec:Entropies}
\begin{defn}[Entropy rate and KL divergence]
The entropy rate of a PFSA $G$ is the entropy rate of the stochastic process $G$ generates~\cite{cover2012elements}. Similarly, the KL divergence of a PFSA $G'$ from the PFSA $G$ is the KL divergence of the process generated by the $G'$ from that of $G$.
More precisely, we have the
\[
    \mathcal{H}(G) = -\lim_{d\rightarrow\infty}\frac{1}{d}\sum_{x\in\Sigma^{d}}Pr(x)\log Pr(x),
\]
and the KL divergence
\[
    \mathcal{D}_{\textrm{KL}}\parenBar{G}{G'} = \lim_{d\rightarrow\infty}\frac{1}{d} \sum_{x\in\Sigma^{d}}Pr_{G}(x)\log\frac{Pr_{G}(x)}{Pr_{G'}(x)},
\]
whenever the limits exist.
\end{defn}

\begin{thm}[Closed-form Formula for Entropy Rate]
The entropy rate of a PFSA $G = \paren{\Sigma, Q, \delta, \pitilde}$ is given by 
\[
    \mathcal{H}(G) = \sum_{q\in{Q}}\mathbf{p}_{G}(q)\cdot h(\pitilde(q)),
\]
where $h(\cdot)$ is the entropy of a probability distribution.
\end{thm}
\vspace{-.2cm}
\textit{proof omitted.}
\vspace{.1cm}

\begin{thm}[Closed-form Formula for KL Divergence]
Let $G = \paren{\Sigma, Q, \delta, \pitilde}$ and $G' = \paren{\Sigma, Q', \delta', \pitilde'}$ be two PFSA, and let $\mathbf{p}_{G}(q, q')$ be the joint $G$-probability ofjoint state $(q, q')$
\footnote{The formal definition of joint $G$-probability needs long and technical derivation, which is outside the main focus of this paper. We can interpret $\mathbf{p}_{G}(q, q')$ as follows. Suppose we have a sample path generated by $G$, and we run the sample path on both $G$ and $G'$ (from arbitrary initial states) and calculate the frequency of the event ``$G$ is in state $q$ and $G'$ is in state $q'$'' as a function of sequence length $d$. The frequency can be shown to converges as $d$ approaches infinity and the limit is $\mathbf{p}_{G}(q, q')$.}, 
then we have the KL divergence of $G'$ from $G$ is given by 
\[
    \sum_{(q, q')\in Q\times{Q'}}p_{G}(q, q')D_{\textrm{KL}}\parenBar{\pitilde(q)}{\pitilde(q')},
\]
where $D_{\textrm{KL}}\parenBar{\cdot}{\cdot}$ is the KL divergence between two probability distributions.
\end{thm}
\vspace{-.2cm}
\textit{proof omitted.}
\vspace{.1cm}

\subsection{Log-likelihood}
\label{subsec:log-likelihoodConvergence}
\begin{defn}[Log-likelihood]
Let $x\in\Sigma^{d}$, the log-likelihood~\cite{cover2012elements} of a PFSA $G$ generating $x$ is given by
\[
    L(x, G) = -\frac{1}{d}\log Pr_{G}(x).
\]
\end{defn}

\begin{thm}[Convergence of Log-likelihood]
Let $G$ and $G'$ be two irreducible PFSA, and let $x\in\Sigma^d$ be a sequence generated by $G$. Then we have
\[
    L(x, G')\rightarrow \mathcal{H}(G) + \mathcal{D}_{\textrm{KL}}\parenBar{G}{G'},
\]
in probability as $d\rightarrow\infty$.
\end{thm}
\begin{IEEEproof}[sketch of proof]
We first notice that
\begin{align*}
    &\sum_{x\in\Sigma^{d}}Pr_{G}(x)\log\frac{Pr_{G}(x)}{Pr_{G'}(x)}\\
    =& \sum_{x\in\Sigma^{d-1}}\sum_{\sigma\in\Sigma}Pr_{G}(x)\mathbf{p}_{G}(x)\left.\Pitilde_{G}\right|_{\sigma}\log\frac{Pr_{G}(x)\mathbf{p}_{G}(x)\left.\Pitilde_{G}\right|_{\sigma}}{Pr_{H}(x)\mathbf{p}_{G'}(x)\left.\Pitilde_{G'}\right|_{\sigma}}\\
    =&\sum_{x\in\Sigma^{d-1}}Pr_{G}(x)\log\frac{Pr_{G}(x)}{Pr_{H}(x)}\\
    &+ \underbrace{\sum_{x\in\Sigma^{d-1}}Pr_{G}(x)\sum_{\sigma\in\Sigma}\mathbf{p}_{G}(x)\left.\Pitilde_{G}\right|_{\sigma} \log\frac{\mathbf{p}_{G}(x)\left.\Pitilde_{G}\right|_{\sigma}}{\mathbf{p}_{G'}(x)\left.\Pitilde_{G'}\right|_{\sigma}}}_{D^{d}}.
\end{align*}
By induction, we have $\mathcal{D}_{\textrm{KL}}\parenBar{G}{G'} = \lim_{d\rightarrow\infty}\frac{1}{d}\sum_{i=1}^{d}D^{i}$, 
and hence by Ces\`{a}ro summation theorem, $\mathcal{D}_{\textrm{KL}}\parenBar{G}{G'} = \lim_{d\rightarrow\infty} D^{d}$ whenever the limit exists.

Let $x=\sigma_1\sigma_2...\sigma_n$ be a sequence generated by $G$. Let $x^{[i]}$ be the truncation of $x$ at the $i$-th symbols, we have
\begin{align*}
    &-\frac{1}{n}\sum_{i=1}^{n}\log\mathbf{p}_{G'}\paren{x^{[i-1]}}\left.\Pitilde_{G'}\right|_{\sigma_i}\\
    =& \underbrace{\frac{1}{n}\sum_{i=1}^{n}\log\frac{\mathbf{p}_{G}\paren{x^{[i-1]}}\left.\Pitilde_{G}\right|_{\sigma_{i}}}{\mathbf{p}_{G'}\paren{x^{[i-1]}}\left.\Pitilde_{G'}\right|_{\sigma_i}}}_{A_{x, n}}\underbrace{-\frac{1}{n}\sum_{i=1}^{n}\log\mathbf{p}_{G}\paren{x^{[i-1]}}\left.\Pitilde_{G}\right|_{\sigma_i}}_{B_{x, n}}.
\end{align*}
Since the stochastic process $G$ generates is ergodic, we have
\[
    \lim_{n\rightarrow\infty}A_{x,n} = \lim_{d\rightarrow\infty}D^{d} = \mathcal{D}_{\textrm{KL}}\parenBar{G}{G'},
\]
and $\lim_{n\rightarrow\infty}B_{x,n} = \mathcal{H}(G)$.
\end{IEEEproof}
\begin{exmpl}
In this example we show the convergence of log-likelihood using the PFSA $G$ on the left of Fig.~\ref{fig:M2_with_q_lambda_and_M2} and the PFSA $H$ that is the induced subgraph on $q_{00}, q_{01}, q_{10}$, and $q_{11}$ in Fig.~\ref{fig:M4}. We have
\begin{align*}
    \mathcal{H}(G)\approx 0.9710,\quad
    \mathcal{H}(H) \approx 0.8069,
\end{align*}
and
\begin{align*}
    \mathcal{D}_{\textrm{KL}}\parenBar{G}{H} \approx 0.2266,\quad \mathcal{D}_{\textrm{KL}}\parenBar{H}{G} \approx 0.2030.
\end{align*}
Let us use $G\rightarrow{x}$ as the short hand for $x$ is generated by $G$, we show in Fig.~\ref{fig:log-likelihoodConverge} the log-likelihood of $G$ producing a sequence $x$ generated by $G$ (top left), the log-likelihood of $H$ producing a sequence $x$ generated by $H$ (top right), the log-likelihood of $H$ producing a sequence $x$ generated by $G$ (bottom left), and the log-likelihood of $G$ producing a sequence $x$ generated by $H$ (bottom right). We can clear see that the convergence of log-likelihood from the plots.

\begin{figure}[t]
    \centering
    \begin{tikzpicture}[scale=.44, every node/.style={transform shape}]
        \def\w{5.1}
        \def\h{3.6}
        \node[label={[rotate=90, yshift=.2cm, xshift=2.4cm]left:{\Huge log-likelihood}}] at (-\w, \h) {\includegraphics[scale=1., trim = 0 0 10 0, clip]{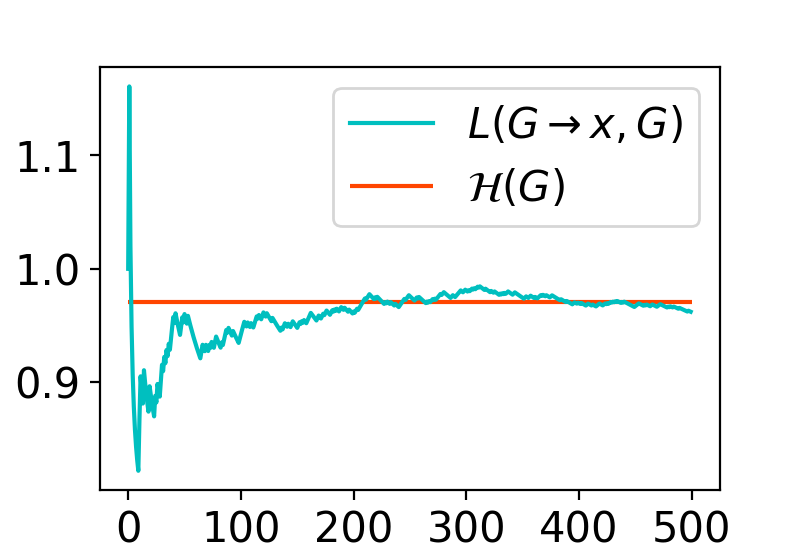}};
        \node at (\w, \h) {\includegraphics[scale=1., trim = 0 0 10 0, clip]{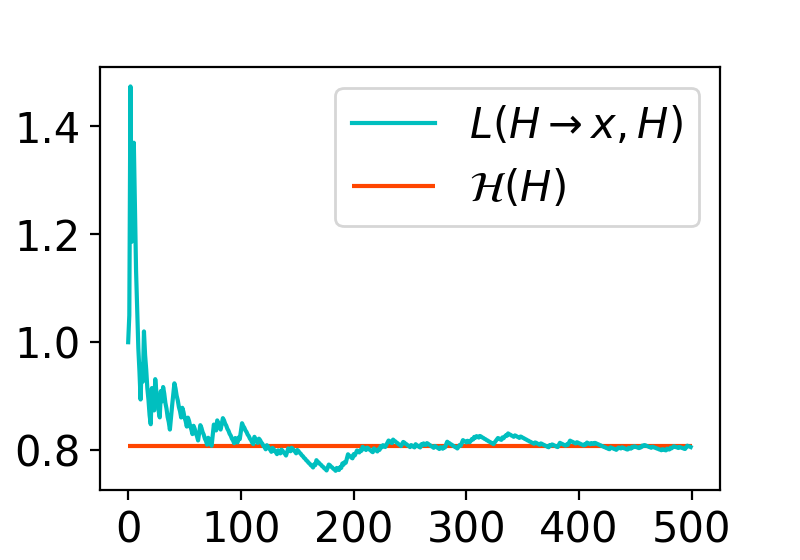}};
        \node[
            label={[rotate=90, yshift=.2cm, xshift=2.4cm]left:{\Huge log-likelihood}},
            label={[xshift=.5cm]below:{\Huge sequence length}}
        ] at (-\w, -\h) {\includegraphics[scale=1., trim = 0 0 10 0, clip]{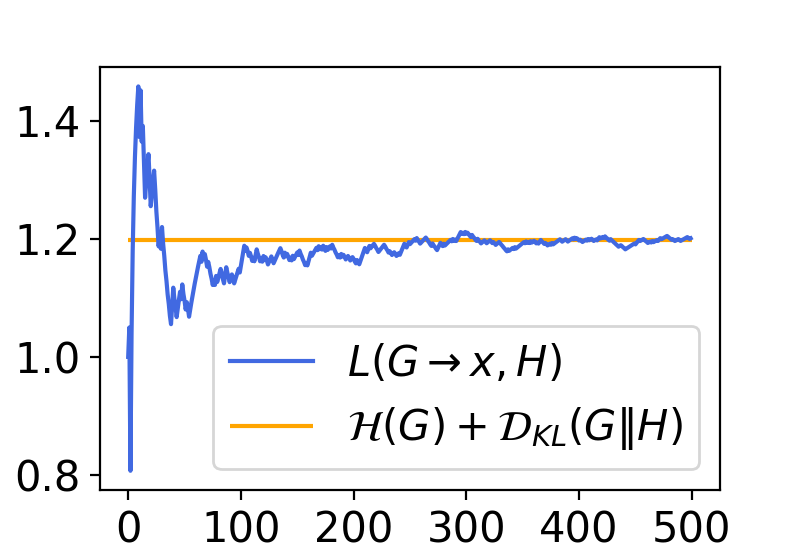}};
        \node[label={[xshift=.5cm]below:{\Huge sequence length}}] at (\w, -\h) {\includegraphics[scale=1., trim = 0 0 10 0, clip]{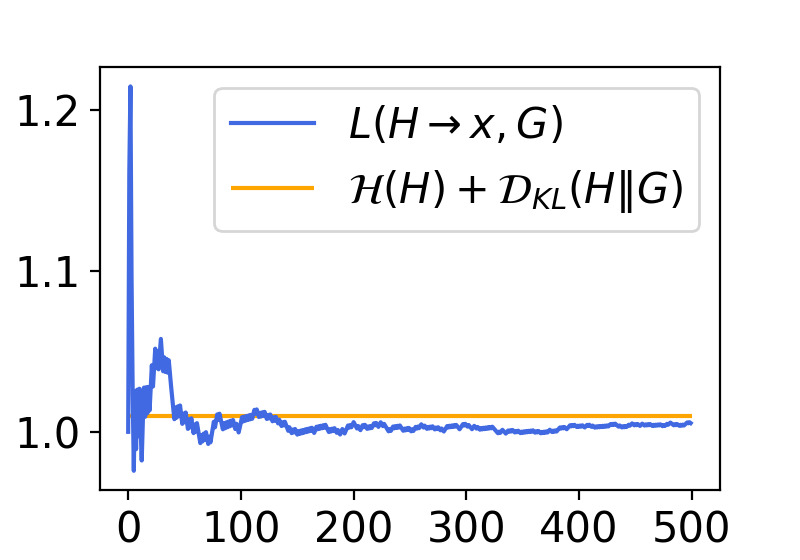}};
    \end{tikzpicture}
    \caption{Examples of log-likelihood convergence. The horizontal lines are the limits calculated by the closed-form formulae of entropy rates and KL divergences.}
    \label{fig:log-likelihoodConverge}
\end{figure}

\end{exmpl}

\section{\textsf{Smash2.0}}
\label{sec:Smash2.0}
With the assumption of discrete-valued input, we first show in Sec.~\ref{subsec:Smash2.0} how to use log-likelihood convergence to define a pairwise distance between sequences. Because PFSA is a model for sequences on finite alphabet, continuous-valued input should first be quantized to discrete ones before being modeled by PFSA. So we discuss in Sec.~\ref{subsec:Quantization} ways of doing quantization and how their fitness can be evaluated. In Sec.~\ref{subsec:Compare_SmashsAndFastDTW}, we compare \textsf{Smash2.0} to \textsf{Smash} and \text{fastDTW} in both performance and efficiency.

\subsection{\textsf{Smash2.0}: Distance between Time Series}
\label{subsec:Smash2.0}
The way we calculate distance between two sequences is as follows. We first choose a set of PFSA $\mathcal{G} = \set{G_0,\dots, G_k}$ as base, and the coordinate for a sequence $x$ is defined to be 
\[
    \paren{L\paren{x, G_0}, \dots, L\paren{x, G_k}}, 
\]
where $L(x, G)$ is the log-likelihood of $G$ generating $x$, as defined in Sec.~\ref{subsec:log-likelihoodConvergence}. The distance between a pair of sequences can then be any valid distance between their coordinates. 

For example, for the two numerical experiments in Sec.\ref{sec:Application}, we use $\mathcal{G}$ that contains four PFSA
\begin{eqnarray*}
    0. &\delta\paren{q_0, 0} = q_0, \delta\paren{q_0, 1} = q_1, \delta\paren{q_1, 0} = q_0, \delta\paren{q_1, 1} = q_1,&\\
    &\pitilde\paren{q_-1}=(.3, .7), \pitilde\paren{q_1}=(.7, .3),&\\
    1. &\delta\paren{q_0, 0} = q_0, \delta\paren{q_0, 1} = q_1, \delta\paren{q_1, 0} = q_1, \delta\paren{q_1, 1} = q_0,&\\
    & \pitilde\paren{q_-1}=(.3, .7), \pitilde\paren{q_1}=(.7, .3),&\\
    2. &\delta\paren{q_0, 0} = q_1, \delta\paren{q_0, 1} = q_2, \delta\paren{q_1, 0} = q_2, \delta\paren{q_1, 1} = q_0,&\\
    &\delta\paren{q_1, 0} = q_0, \delta\paren{q_2, 1} = q_1,& \\
    & \pitilde\paren{q_-1}=(.3, .7), \pitilde\paren{q_1}=(.7, .3), \pitilde\paren{q_2}=(.6, .4),&\\
    3. &\delta\paren{q_0, 0} = q_0, \delta\paren{q_0, 1} = q_1, \delta\paren{q_1, 0} = q_2, \delta\paren{q_1, 1} = q_3,&\\
    &\delta\paren{q_1, 0} = q_0, \delta\paren{q_2, 1} = q_1, \delta\paren{q_3, 0} = q_2, \delta\paren{q_3, 1} = q_3,&\\
    & \pitilde\paren{q_-1}=(.3, .7), \pitilde\paren{q_1}=(.7, .3),& \\
    & \pitilde\paren{q_1}=(.8, .2), \pitilde\paren{q_3}=(.2, .8).& 
\end{eqnarray*}
and the distance between coordinates to be the total variation distance ($l_0$ distance). One thing we'd like to point out is that, we use these four PFSA as base primarily for simplicity. A better way of forming $\mathcal{G}$, especially for supervised problems, is to use PFSA inferred by, for example, \algo~proposed in Sec.~\ref{sec:inference_PFSA}, from the training set.

\subsection{Quantization of Continuous Sequence}
\label{subsec:Quantization}
The simplest approach to turn a continuous sequence to a symbolic one with alphabet size $k$ is by choosing $k-1$ cut-off points $p_1 < p_2 < \cdots <p_{k-1}$. With the additional assumption that $p_0 = -\infty$ and $p_k = +\infty$, we can replace a data point $p$ in the continuous sequence with symbol $i$ if $p\in[p_i  p_{i+1})$. We call the set of cut-off points a \emph{partition}. The most common practice of choosing a partition is to apply the entropy maximization principle, in which the $p_i$s are chosen so that we have as equal as possible numbers of data points in each interval. However, we can also perturb the cut-off points a little bit so that the distribution of symbols in the quantized sequences have smaller entropy. 

We can foresee the simple quantization above may work sub-optimally in the following two scenarios: first, the input sequences share a common trend; second, the input sequences have shifts and re-scalings. An example of the first case is the stock prices dataset, and examples for the second case are common among datasets of voltages that have different reference point or sound that have difference volumes. In some of these cases, We may want to first take derivatives (or detrend) and normalize the sequences to have empirical mean $0$ and variance $1$ before they are translated to symbolic sequences using a partition.

In our study of PFSA modeling, we try to use as many different quantization schemes as possible in the hope of exploring the dataset to a fuller extent. We do so by combining different decisions for whether to detrend and normalize and different choice of partitions to form a pool of quantization schemes. To record the parameters for a quantization scheme, we develop the following shorthand: 1) $\mathsf{D}\mathsf{d}$, with $\mathsf{d}$ being an non-negative integer, means detrending $d$ times; 2) $\textsf{N}\mathsf{0}$ means not to apply normalization, and $\textsf{N}\mathsf{1}$, apply normalization; 3) $[p_1\,p_2\, \cdots\,p_{k-1}]$ means a $k$-part partition with cut-off points $p_i$s. As an example, $\mathsf{D1N1}[3.]$ means that we first detrend once, then normalize, and finally replace all data points that are less than $3.$ with a symbol $0$, and those greater than or equal to $3.$, symbol $1$.

To evaluate the quality of a quantization scheme, we first need to calculate the distance matrix $D$ on the training set, with $D_{i,j}$ being the distance between the $i$-th sequence and $j$-th sequence. Assuming that there are $n$ sequences and $l_i$ is the class label of the $i$-th sequence, we define the \emph{average inter-class distances} to be
\[
    s(D) = \frac{\sum_{i= 1}^{n}\sum_{j= 1}^{n}\delta_{l_il_j}D_{i,j}}{\sum_{i= 1}^{n}\sum_{j= 1}^{n}\delta_{l_il_j}},
\]
and \emph{average intra-class distance} to be
\[
    d(D) = \frac{\sum_{i= 1}^{n}\sum_{j= 1}^{n}\paren{1 - \delta_{l_il_j}}D_{i,j}}{\sum_{i= 1}^{n}\sum_{j= 1}^{n}\paren{1 - \delta_{l_il_j}}},
\]
where $\delta_{ab} = 1$ if $a = b$ and $0$ if otherwise. Now we can evaluate the quality of the quantization scheme by the ratio $r(D) = s(D)/d(D)$. The intuition behind this quality measurement is that a good quantization scheme should make the average distances between sequences from the same class small and the average distances between sequences from different classes large.

\subsection{\textsf{Smash}, \textsf{Smash2.0}, and \textsf{fastDTW}}
\label{subsec:Compare_SmashsAndFastDTW}
\begin{figure}[t]
\centering
\includegraphics[scale=1]{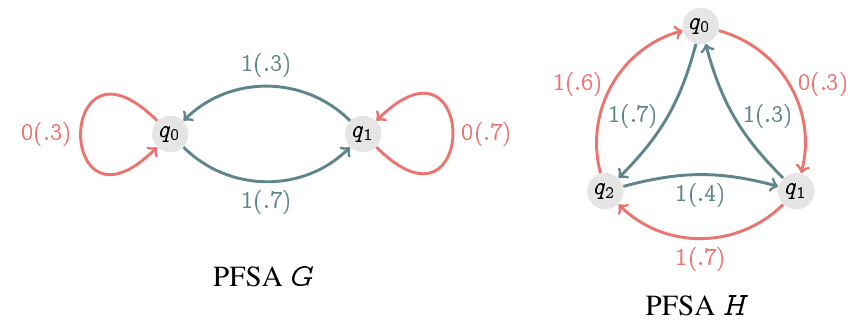}
\caption{PFSA that generate sequences for the comparison Sec.~\ref{subsec:Compare_SmashsAndFastDTW}.}
\label{fig:AlgComp_SeqGenerator}
\end{figure}

\begin{figure}[t]
\centering
\includegraphics[scale=1.]{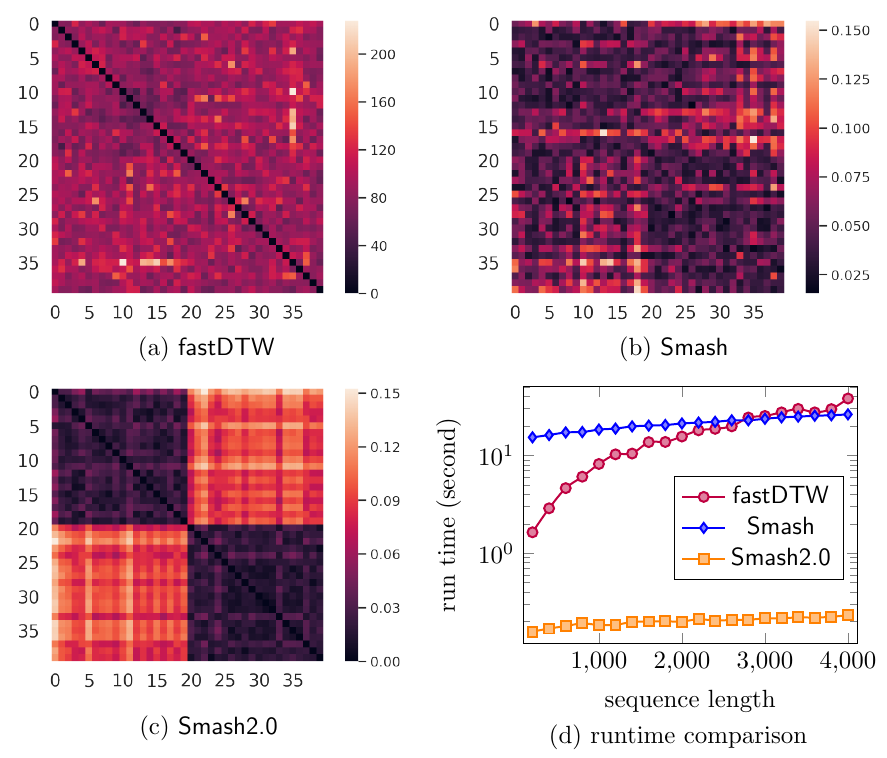}
\caption{Figure (a-c) are heatmaps of the distance matrices calculated by \textsf{fastDTW}, \textsf{Smash}, and \textsf{Smash2.0}. We can see that although the two PFSA generate drastically different stochastic processes, \textsf{fastDTW} fails to capture the distinction. \textsf{Smash} does a slightly better job than \textsf{fastDTW}, but \textsf{Smash2.0} does much better. We show in (d) that \textsf{Smash2.0} runs much faster than the other two algorithms.}
\label{fig:AlgComp}
\end{figure}

In this section we compare three similarity measurement of time series, \textsf{fastDTW} proposed in \cite{salvador2004fastdtw}, \textsf{Smash} proposed in \cite{chattopadhyay2014data}, and \textsf{Smash2.0}. We show using a synthetic dataset that \textsf{Smash2.0} outperforms the other two algorithms in both performance and efficiency. The dataset we use contains two classes, each with $20$ sequences of length $500$ generated by the two PFSA, $G$ and $H$, as shown in Fig.~\ref{fig:AlgComp_SeqGenerator}. The two PFSA are functionally different as shown by their their KL divergence. We have $\mathcal{D}_{\textrm{KL}}\parenBar{G}{H} \approx 0.19677$ and $\mathcal{D}_{\textrm{KL}}\parenBar{H}{G} \approx 0.20756$. For \textsf{fastDTW}, we use the Python package \textsf{fastDTW} (\url{https://pypi.org/project/fastdtw/}) with default value $1$ for radius. We use $2$ for number of reruns for \textsf{Smash} because of its probabilistic nature. In Fig.~\ref{fig:AlgComp}(a-c) we show the heatmaps of the distance matrices given by the three algorithms. In Fig.~\ref{fig:AlgComp} (d), we show running time of the three algorithms on dataset constructed the same way as above but with sequence length ranging from $200$ to $4000$, with $200$ increment. The computer we used to do the calculation has Intel E5-2680v4 2.4GHz CPU and 64GB memory. 

\section{Applications}
\label{sec:Application}
In this section, we use \textsf{Smash2.0} to study two real world problems. 
\subsection{Dataset 1: Motor Movement Imagery Dataset}
\label{dataset:EEG}
This dataset is an excerpt of a dataset from  PysioNet\cite{goldberger2000physiobank}\footnote{\url{http://www.physionet.org/pn4/eegmmidb/}}. The dataset contains 64-channel $160$Hz EEG recorded by the BCI2000 system\cite{schalk2004bci2000}\footnote{\url{http://www.bci2000.org}} while subjects performed different motor imagery tasks. There are four tasks in the dataset and we focus on two of them: 
\begin{itemize}
    \item[\textsf{TM}:] A target appears on either the left or the right side of the screen. The subject opens and closes the corresponding fist until the target disappears. Then the subject relaxes.
    \item[\textsf{TI}:] The same as the first task, except the subject \emph{imagines} opening and closing the corresponding fist but doesn't really move.
\end{itemize}

\begin{figure}[t]
    \centering
    \begin{tikzpicture}[scale=.82, every node/.style={transform shape}]
        \def\h{.7in}
        \begin{scope}
            \node (TM_title) at (0, \h) {Subject \textsf{S004} task \textsf{TM} Recordings $1$ and $2$};
            \node[
                label={[align=center, rotate=90, yshift=.2cm, xshift=1cm]left:{calibrated\\voltage}},
                label={[yshift=.4cm]below:{time stamp (1/160 second)}}
            ] (TM) at (0, 0) {\includegraphics[scale=1., trim=5 0 0 0, clip]{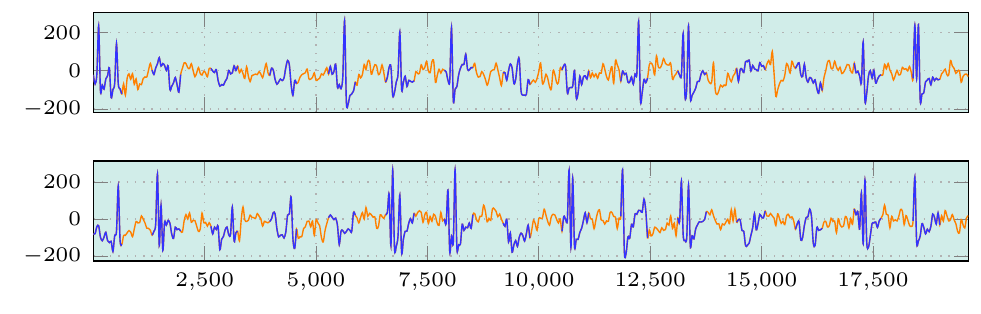}};
        \end{scope}
        \begin{scope}[yshift=-1.5in]
            \node (TI_title) at (0, \h) {Subject \textsf{S004} task \textsf{TI} Recordings $1$ and $2$};
            \node[
                label={[align=center, rotate=90, yshift=.2cm, xshift=1cm]left:{calibrated\\voltage}},
                label={[yshift=.4cm]below:{time stamp (1/160 second)}}
            ] (TI) at (0, 0) {\includegraphics[scale=1., trim=5 0 0 0, clip]{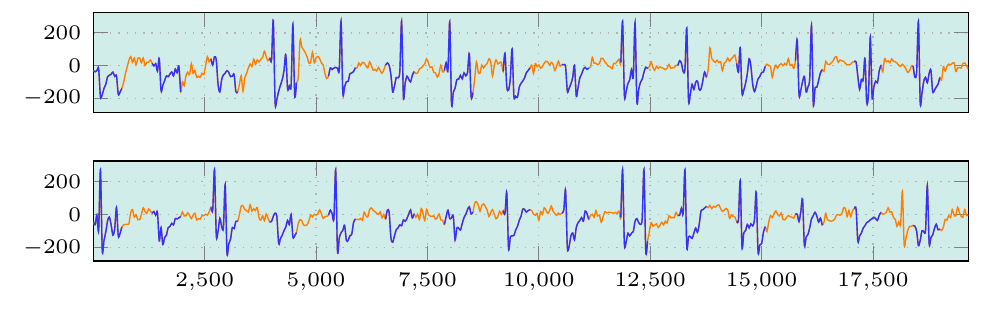}};
        \end{scope}
    \end{tikzpicture}
    \caption{EEG recordings of subject \textsf{S004}.}
    \label{fig:S004_EEG}
\end{figure}

\begin{figure}[t]
    \centering
    \begin{tikzpicture}[scale=.82, every node/.style={transform shape}]
        \def\h{.7in}
        \begin{scope}
            \node (TM_title) at (0, \h) {Subject \textsf{S001} task \textsf{TM} Recordings $1$ and $2$};
            \node[
                label={[align=center, rotate=90, yshift=.2cm, xshift=1cm]left:{calibrated\\voltage}},
                label={[yshift=.4cm]below:{time stamp (1/160 second)}}
            ] (TM) at (0, 0) {\includegraphics[scale=1., trim=5 0 0 0, clip]{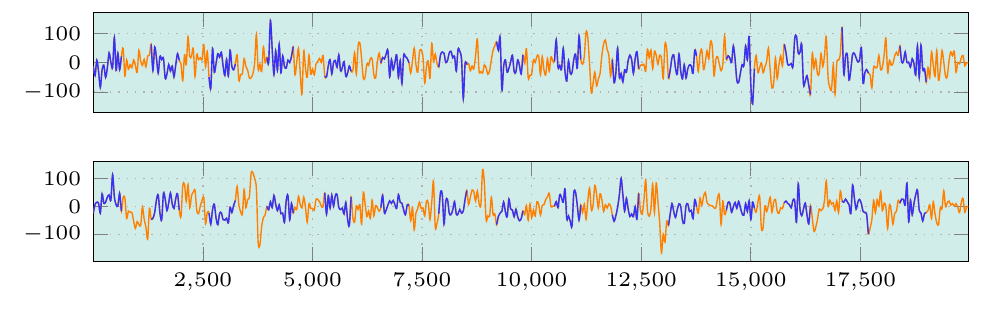}};
        \end{scope}
        \begin{scope}[yshift=-1.5in]
            \node (TI_title) at (0, \h) {Subject \textsf{S001} task \textsf{TI} Recordings $1$ and $2$};
            \node[
                label={[align=center, rotate=90, yshift=.2cm, xshift=1cm]left:{calibrated\\voltage}},
                label={[yshift=.4cm]below:{time stamp (1/160 second)}}
            ] (TI) at (0, 0) {\includegraphics[scale=1., trim=5 0 0 0, clip]{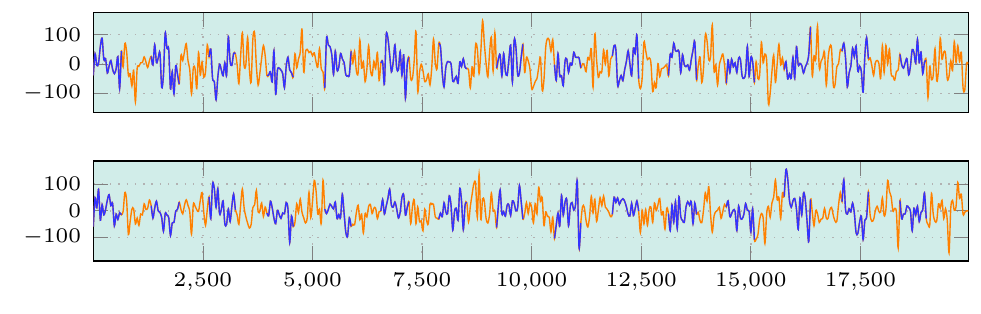}};
        \end{scope}
    \end{tikzpicture}
    \caption{EEG recordings of subject \textsf{S001}.}
    \label{fig:S001_EEG}
\end{figure}

For each subject, three 2-minute EEG recordings are taken for each task and we use the first two recordings to get the results listed below. During each recording, an object appears on the screen for $4$ seconds and disappear for $4$ second, and hence a subject supposedly moves or imagines to move his or her fists for $4$ seconds, then rest for $4$ seconds, and repeat. In Fig.~\ref{fig:S004_EEG} and \ref{fig:S001_EEG}, the top two EEG recordings are for task \textsf{TM} and the bottom two, task \textsf{TI}. We color the rest sections blue, while the movement/imaginary movement sections orange. 

For each subject we form a dataset with $56$ sequence for each task, whose composition is detailed in Tab.~\ref{tab:Application_DecriptionOfMotorDataset}. We drop the first section and the last section from each recordings since they tend to be more noisy.
\begin{table}[ht]
    \centering
    \begin{tabular}{|R{.28in}|R{1.1in}|R{1.6in}|}
        \hline
        seq. & \textsf{TM} dataset & \textsf{TI} dataset \\
        \hline
         $0$-$13$ & rest from rec.~1 & rest from rec.~1 \\
        $14$-$27$ & rest from rec.~2 & rest from rec.~2\\
        $28$-$41$ & movement from rec.~1 & \emph{imaginary} movement from rec.~1\\
        $42$-$55$ & movement from rec.~2 & \emph{imaginary} movement from rec.~2\\ 
        \hline
    \end{tabular}
    \caption{The composition of the \textsf{TM} and \textsf{TI} datasets}
    \label{tab:Application_DecriptionOfMotorDataset}
\end{table}

\begin{figure}[t]
    \centering
    \begin{tikzpicture}[scale=.37, every node/.style={transform shape}]
        \def\w{5.8}
        \node (TM) at (-\w, 0) {\includegraphics[scale=1, trim=70 10 30 20, clip]{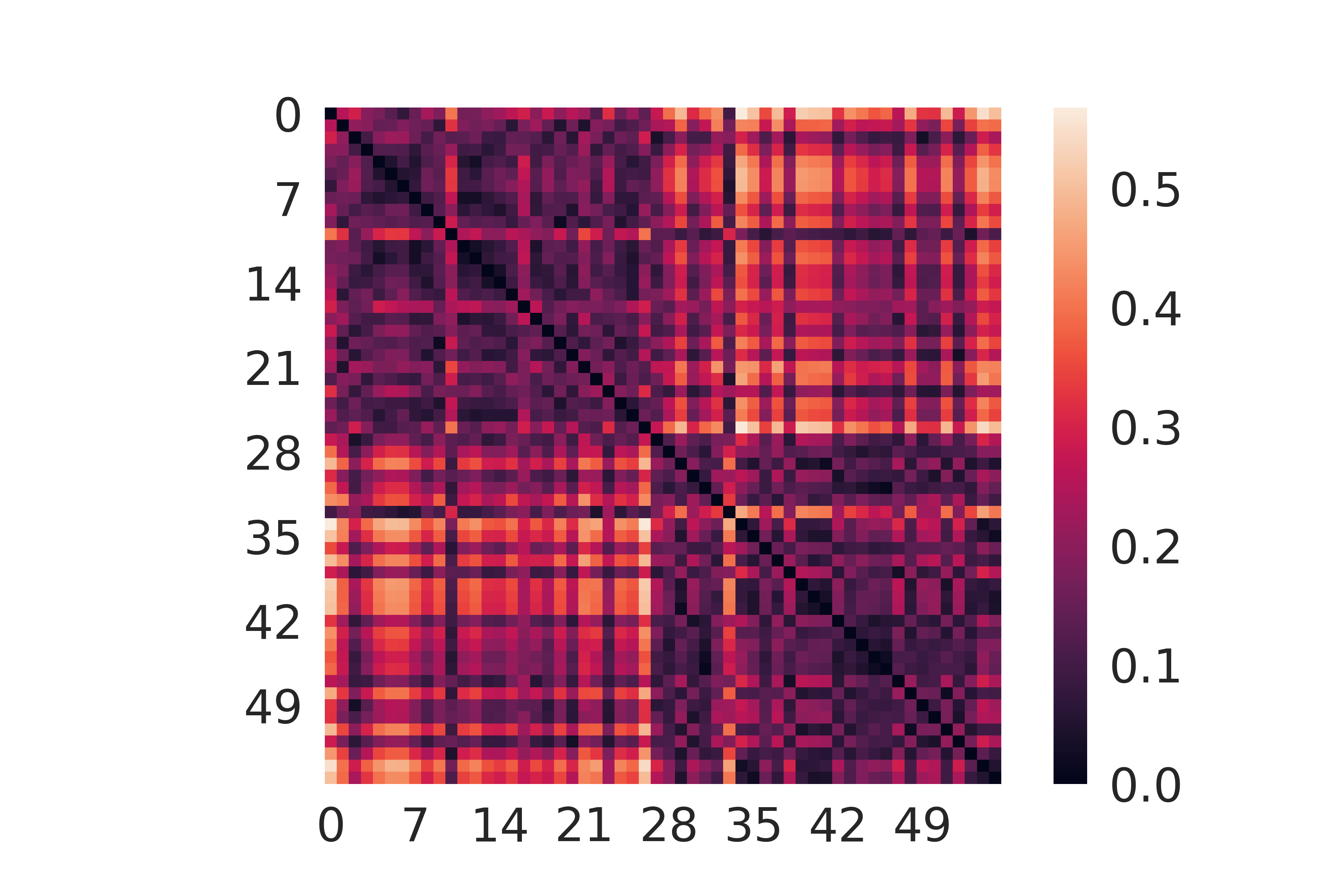}};
        \node (TI) at (\w, 0) {\includegraphics[scale=1, trim=70 10 30 20, clip]{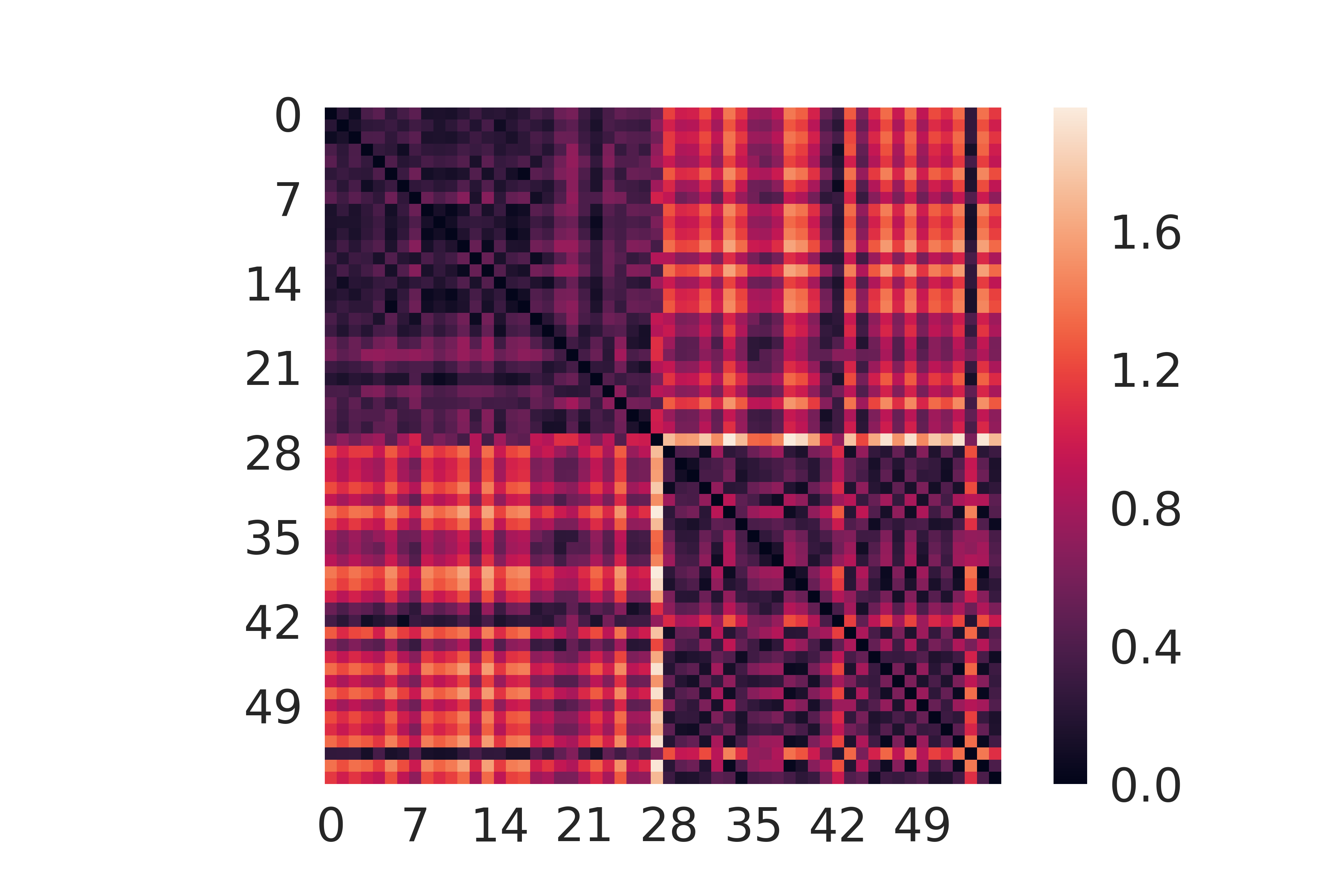}};
    \end{tikzpicture}
    \caption{Heatmaps for \textsf{S004}}
    \label{fig:S004_heatmap}
\end{figure}
The heatmap on the left of Fig.~\ref{fig:S004_heatmap} demonstrates the distance matrix calculated from channel $25$ of the subject \textsf{S004} on task \textsf{TM}. The quantization scheme of the sequences is $\mathsf{D0N1}[-0.4526]$, with $r(D) = .572$. The heatmap on the right of Fig.~\ref{fig:S004_heatmap} demonstrates the distance matrix calculated from channel $21$ of the same subject on task \textsf{TI}. The quantization scheme is $\mathsf{D0N0}[-15.]$, with $r(D) = .451$. We can see that subject \textsf{S004} has drastically different patterns in EEG between rest and (imaginary)movement sections, both from the wave and from the heatmaps of the distance matrices. The rest-movement difference of EEG is persistent across recordings as the distance between both the rest and movement sections from the first and second recordings are relatively insignificant.

\begin{figure}[t]
    \centering
    \begin{tikzpicture}[scale=.37, every node/.style={transform shape}]
        \def\w{5.8}
        \node (TM) at (-\w, 0) {\includegraphics[scale=1, trim=70 10 30 20, clip]{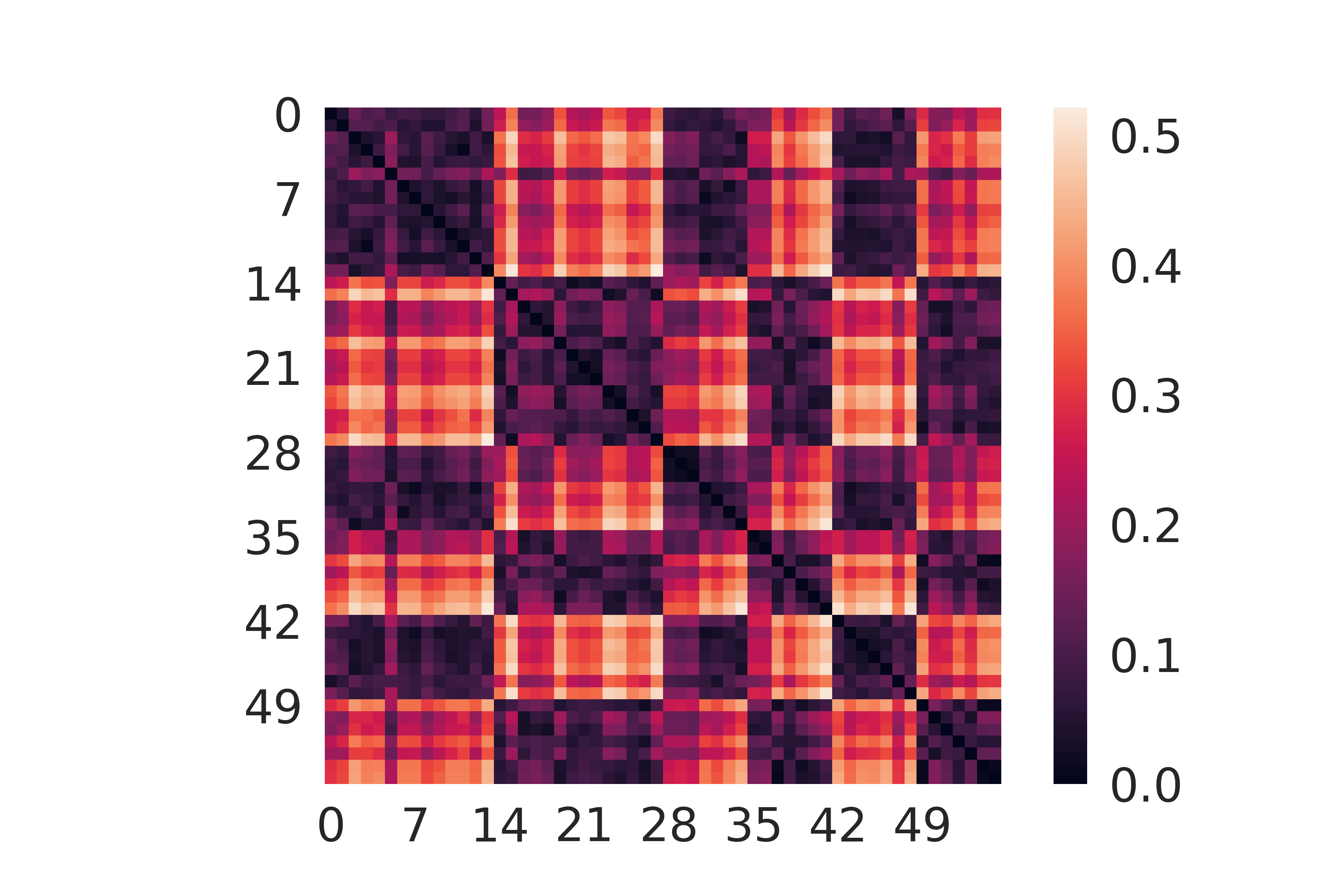}};
        \node (TI) at (\w, 0) {\includegraphics[scale=1, trim=70 10 30 20, clip]{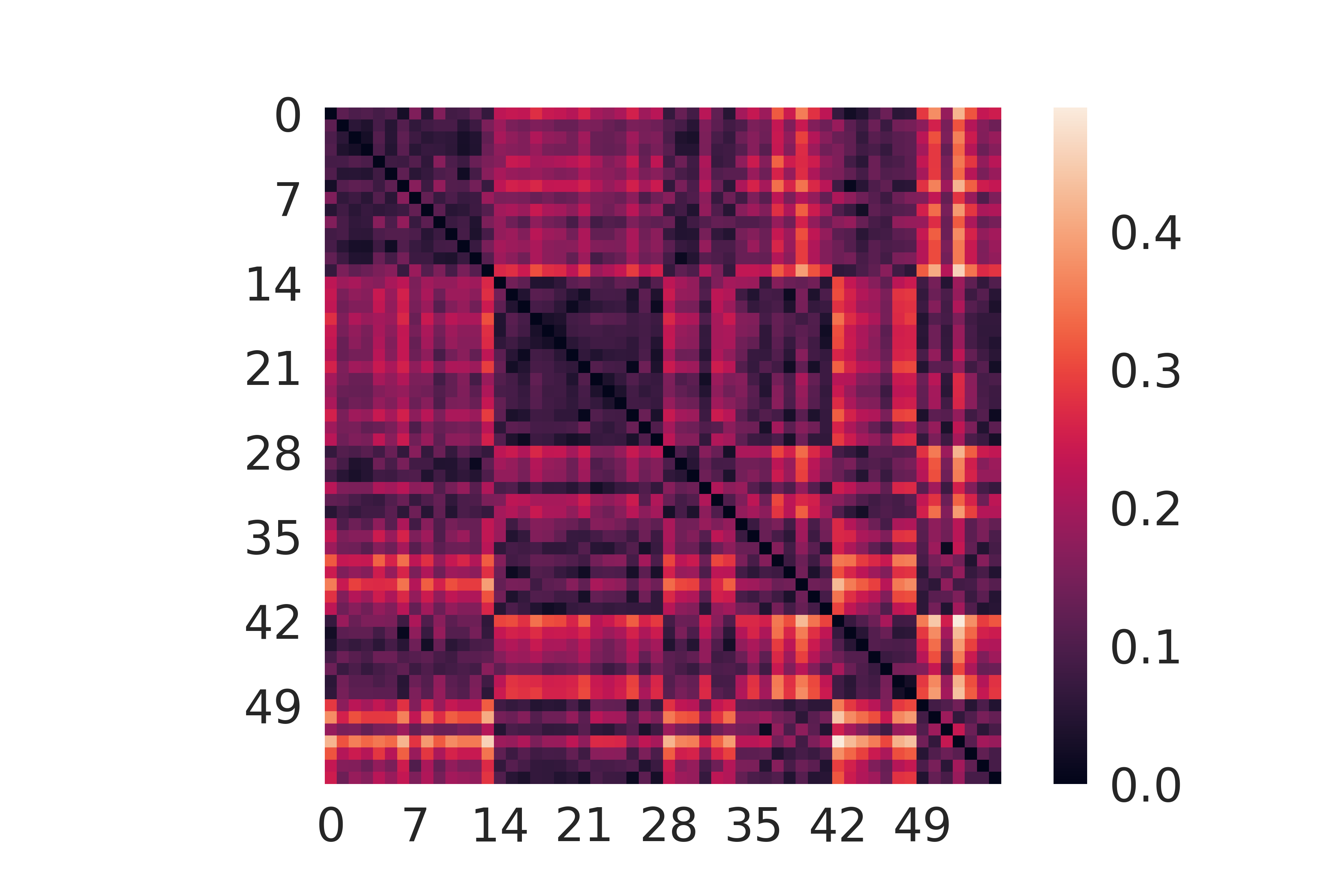}};
    \end{tikzpicture}
    \caption{Heatmaps for \textsf{S001} for tasks \textsf{TM} and \textsf{TI}}
    \label{fig:S001_heatmap}
\end{figure}
The heatmap on the left of Fig.~\ref{fig:S001_heatmap} demonstrates the distance matrix calculated from channel $41$ of subject \textsf{S001} on task \textsf{TM}. The quantization scheme of the sequences is $\mathsf{D0N0}[14.]$ with $r(D) = .825$. The heatmap on the right Fig.~\ref{fig:S001_heatmap} demonstrates the distance matrix calculated from channel $59$ of the same subject on task \textsf{TI}. The quantization scheme is $\mathsf{D1N1}[.4617]$, with $r(D) = .830$. We can see that subject \textsf{S001} does not have significant difference in the EEG between rest and (imaginary) movement sections. Instead the subject seem to be in very different brain states during the two recordings for the same task.

\subsection{Dataset 2: User Identification from Walking Activity}
\label{dataset:WalkingActivity}
This dataset from UCI machine learning repository \cite{Dua:2019} contains 22 subject walking along a predefined trail in the wild. Accelerometer measurements in $x, y, z$ directions were taken with an Android smartphone placed in the chest pocket of each participant. The challenge is to identify a user using his or her pattern of motion. To form a training dataset for each subject, we get $10$ sequence from the beginning of the measurement, each of $500$ time steps long (each time step is about $0.03$ second) and with $250$ time step overlap between two consecutive sequences. There are $12$ participants who has long enough measurement to form training datasets described above. 

\begin{figure}[ht]
    \centering
    \begin{tikzpicture}[scale=.32, every node/.style={transform shape}]
        \def\w{7}
        \node at (-\w, 0) {\includegraphics[scale=1]{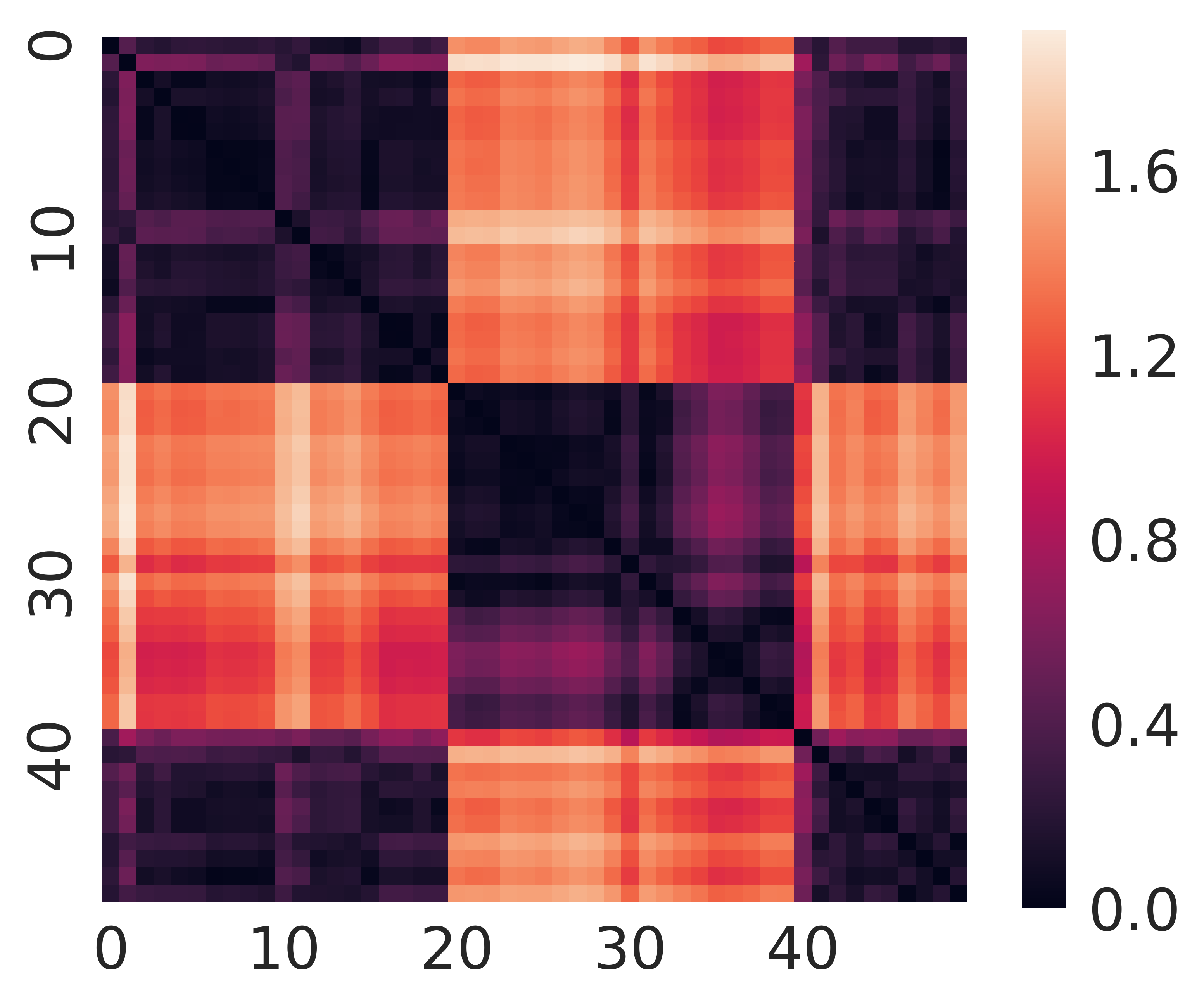}};
        \node at (\w, 0) {\includegraphics[scale=1]{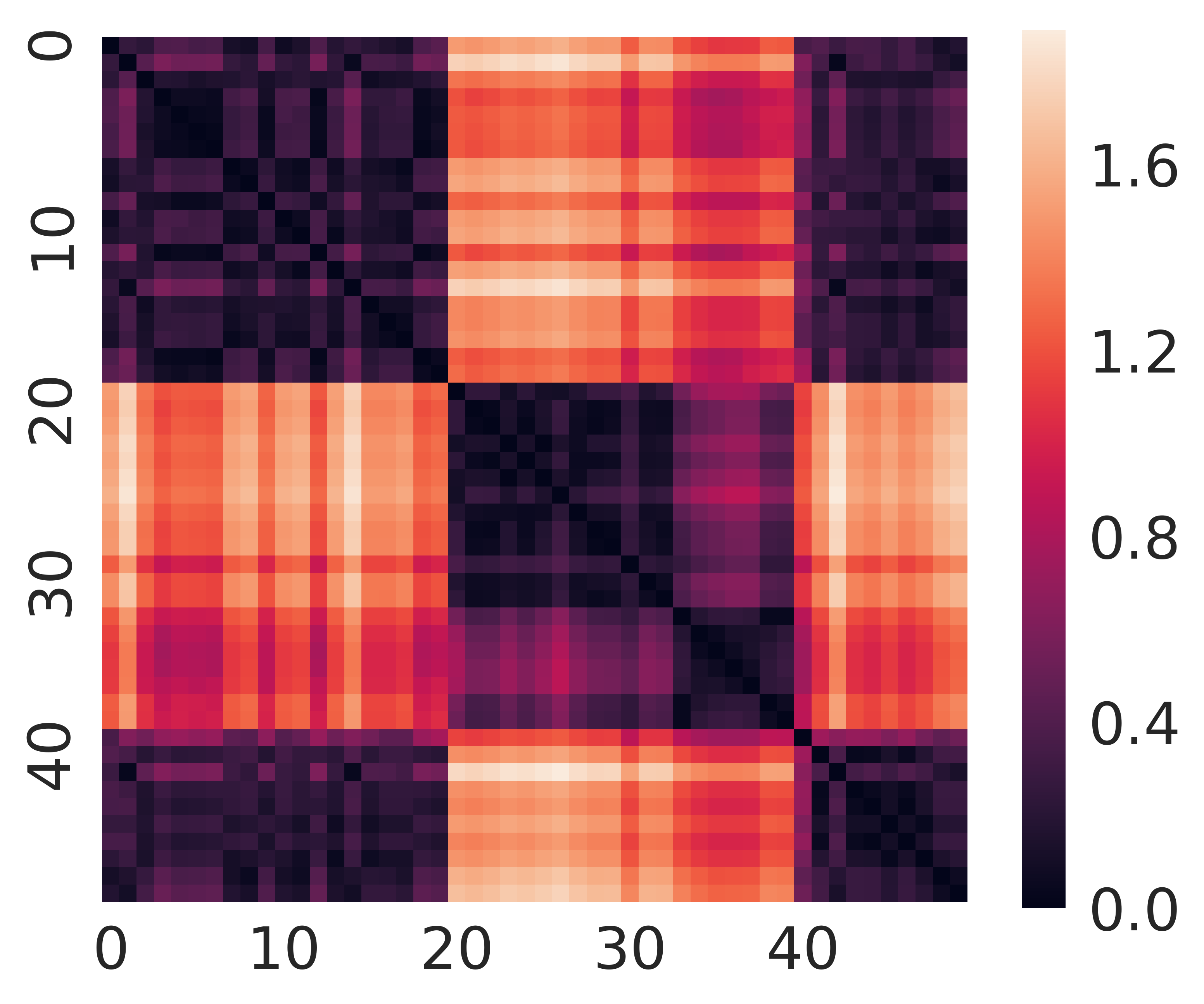}};
    \end{tikzpicture}
    \caption{Heatmaps for best two distances in $x$ direction. On left:  $\mathsf{D0N0}[-1.1169]$ with $r(D)=.230$. On right: $\mathsf{D0N0}[-0.3814]$ with $r(D)=.248$.}
    \label{fig:walker_x}
\end{figure}

\begin{figure}[t]
    \centering
    \begin{tikzpicture}[scale=.32, every node/.style={transform shape}]
        \def\w{7}
        \node at (-\w, 0) {\includegraphics[scale=1]{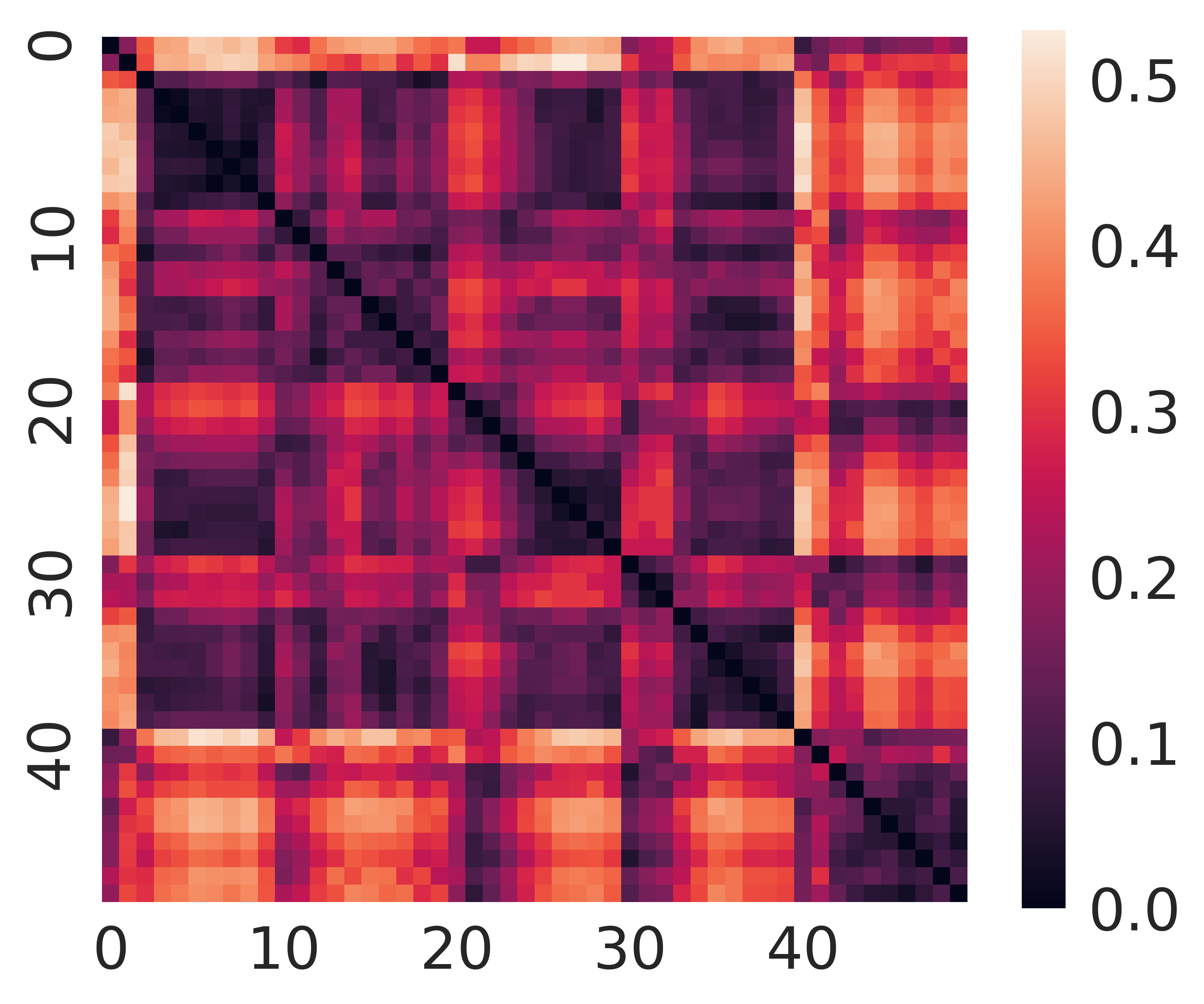}};
        \node at (\w, 0) {\includegraphics[scale=1]{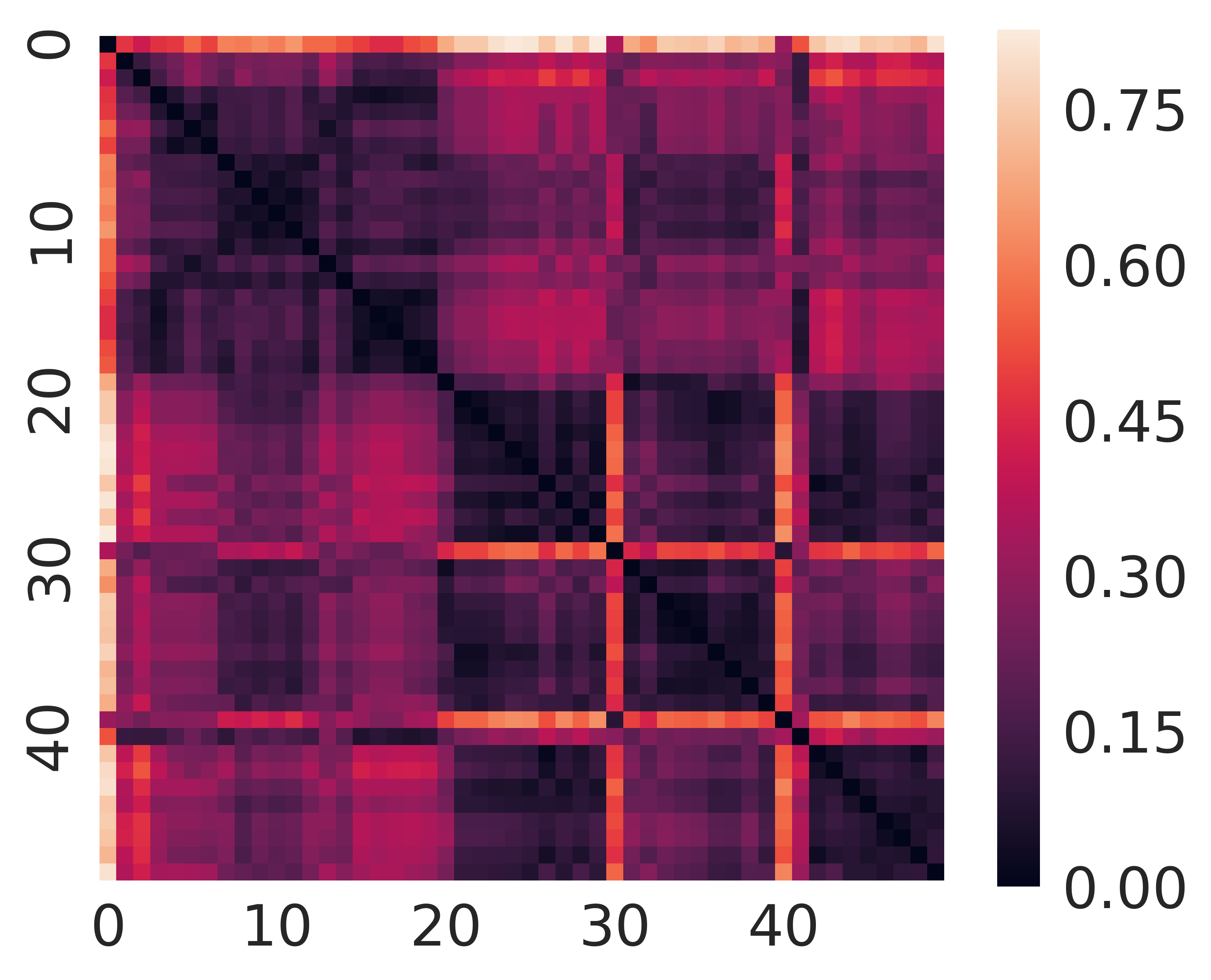}};
    \end{tikzpicture}
    \caption{Heatmaps for best two distances in $y$ direction. On left:  $\mathsf{D1N0}[0.0820]$ with $r(D)=.571$. On right: $\mathsf{D}0{N}1[-0.0438]$ with $r(D)=.577$.}
    \label{fig:walker_y}
\end{figure}

\begin{figure}[ht]
    \centering
    \begin{tikzpicture}[scale=.32, every node/.style={transform shape}]
        \def\w{7}
        \node at (-\w, 0) {\includegraphics[scale=1]{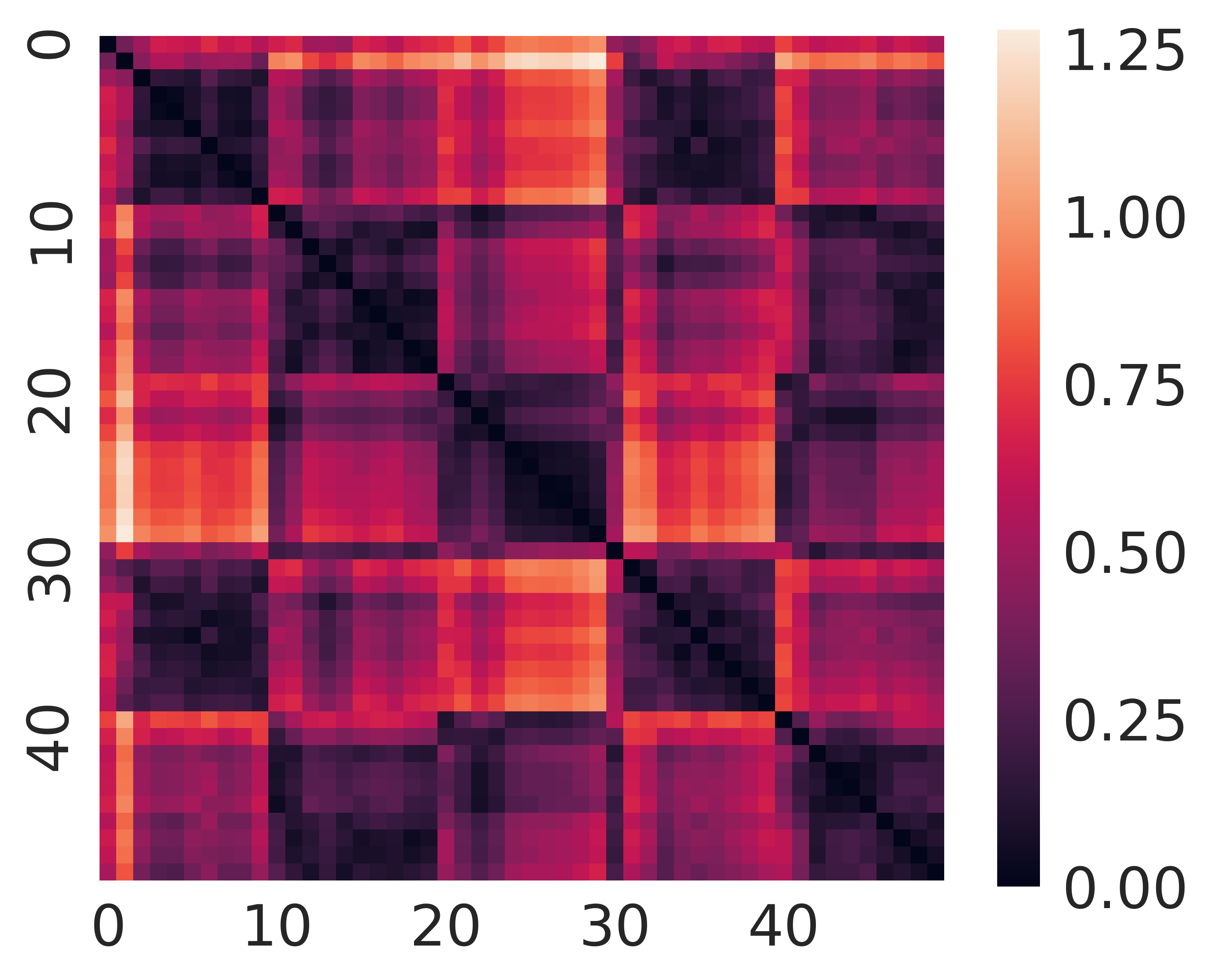}};
        \node at (\w, 0) {\includegraphics[scale=1]{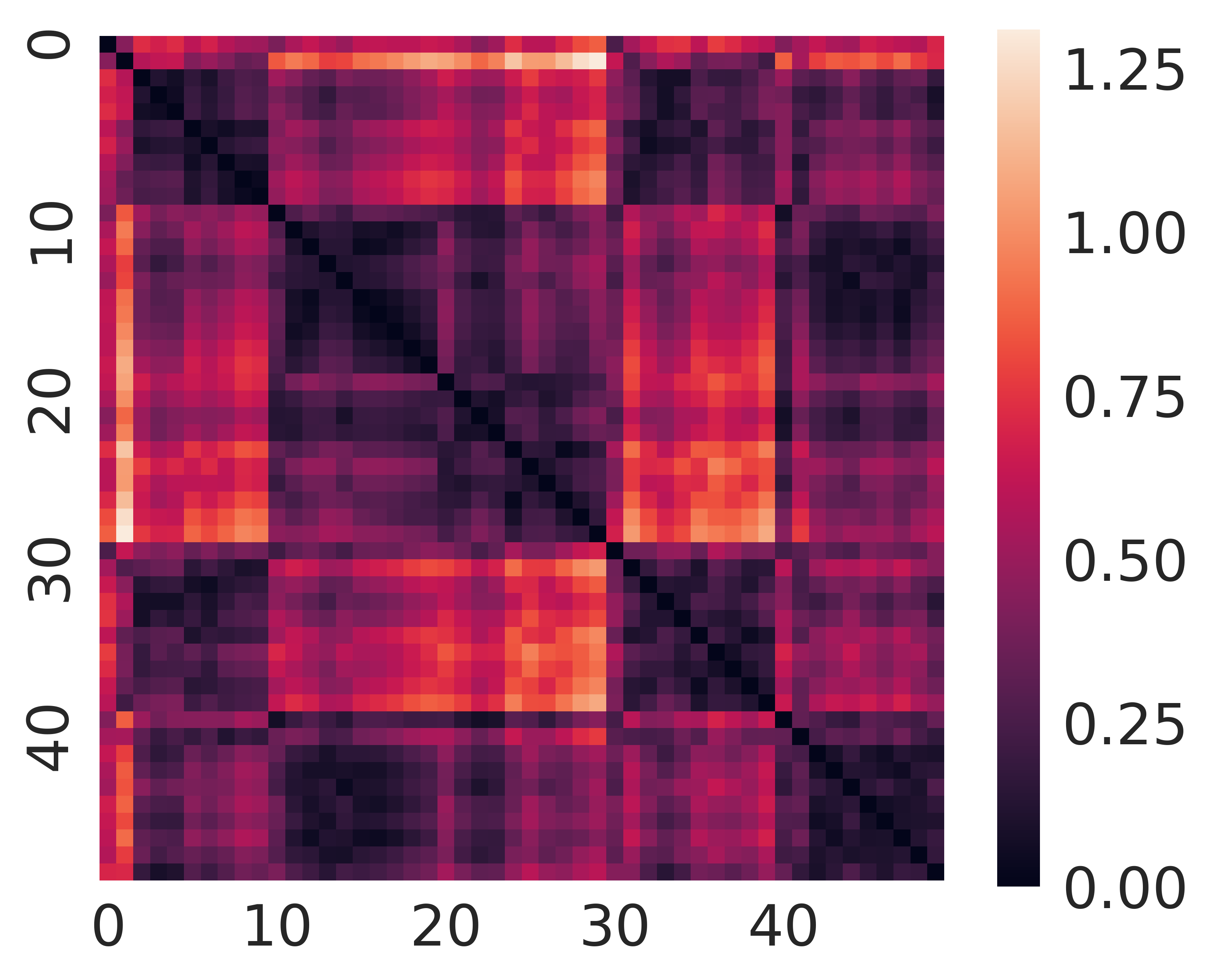}};
    \end{tikzpicture}
    \caption{Heatmaps for best two distances in $z$ direction. On left:  $\mathsf{D0N0}[0.0409]$with $r(D)=.324$. On right: $\mathsf{D0N0}[0.6129]$ with $r(D)=.340$}
    \label{fig:walker_z}
\end{figure}

In Fig.~\ref{fig:walker_x}, \ref{fig:walker_y}, and \ref{fig:walker_z}, we plot the heatmaps of the best two distances for acceleration measurement in the $x$, $y$, and $z$ directions, respectively. Although the distance matrices calculated on all $12$ participants, we demonstrate the heatmaps for the first $5$ participants for clarity.

From this example we can see that although it may be difficult to tell two participant apart by using measurement from one directions, using measurement from multiple channels combined with the freedom of our approach to choose quantization schemes customized for each individual channel, we may boost the classification performance. As for an example, while the distance in the $x$ direction between the first two participants (sequences $0$-$9$, and sequences $10$-$19$) may not be great enough to tell them apart, their patterns of motion in the $z$ direction do have much big distinction as indicated by the the first $20$ by $20$ diagonal block of the two heatmaps in Fig.~\ref{fig:walker_z}.

\section{Conclusion}
\label{sec:Conclusion}
In this paper, we propose a distance metric \textsf{Smash2.0} between time series based on PFSA modeling and sequence likelihood divergence. We give a self-contained introduction to the mathematical foundation of PFSA as a time series model, the quantification of entropy rate and KL divergence of the stochastic process generated by PFSA, and finally, log-likelihood convergence of sample paths. We show how to infer PFSA from sequences and how to evaluate sequence likelihood divergence using log-likelihood convergence. We define the distance metric \textsf{Smash2.0} using sequence likelihood divergence, and with the help of quantization algorithm of continuous data streams, we demonstrate how to apply \textsf{Smash2.0} to the analysis of time series datasets arising from real world scenarios.

Possible future research effort includes 1) finding a better way to choose base PFSA for the \textsf{Smash2.0} in unsupervised settings; 2) finding a base-free way to calculate \textsf{Smash2.0} distance by, for example, inferring a PFSA model from each time series in the dataset.

\bibliographystyle{siam}
\bibliography{bibliography}

\end{document}